\documentclass[10pt,twocolumn,letterpaper]{article}

\usepackage[pagenumbers]{wacv} 

\usepackage{graphicx}
\usepackage{booktabs}
\usepackage{multirow}
\usepackage[table]{xcolor}
\usepackage[most]{tcolorbox}
\usepackage{adjustbox}

\definecolor{UrbanBG}{RGB}{242,247,252}
\definecolor{JasperBG}{RGB}{248,248,248}
\definecolor{StonewallBG}{RGB}{250,247,241}
\definecolor{AluniteBG}{RGB}{250,247,241}
\definecolor{SamsonBG}{RGB}{242,247,252}

\definecolor{UrbanLine}{RGB}{93,139,181}
\definecolor{JasperLine}{RGB}{120,120,120}
\definecolor{StonewallLine}{RGB}{174,137,79}
\definecolor{SamsonLine}{RGB}{93,139,181}
\definecolor{AluniteLine}{RGB}{174,137,79}

\definecolor{WinsBG}{RGB}{232,246,235}
\definecolor{BestBG}{RGB}{232,246,235}

\newcommand{\datasetsep}{%
  \hspace{2pt}%
  {\color{black!30}\vrule width 0.4pt}%
  \hspace{2pt}%
}

\newcommand{\datasetrow}[5]{%
\begin{tcolorbox}[
    enhanced,
    width=\columnwidth,
    colback=white,
    colframe=#2,
    boxrule=0.9pt,
    arc=1.8mm,
    outer arc=1.8mm,
    left=1.5mm,
    right=1.5mm,
    top=2.3mm,
    bottom=1.2mm,
    before skip=0mm,
    after skip=0.5mm,
    title={#1},
    fonttitle=\sffamily\bfseries\footnotesize,
    coltitle=#2,
    attach boxed title to top center={
        yshift=-2.4mm
    },
    boxed title style={
        colback=white,
        colframe=white,
        boxrule=0pt,
        arc=0pt,
        left=2mm,
        right=2mm,
        top=0pt,
        bottom=0pt
    }
]
    \centering

    \makebox[\linewidth][c]{%
        \resizebox{0.24\linewidth}{1.90cm}{%
            \includegraphics{#3}%
        }%
        \hfill
        \resizebox{0.24\linewidth}{1.90cm}{%
            \includegraphics{#4}%
        }%
        \hfill
        \resizebox{0.41\linewidth}{2.00cm}{%
            \includegraphics{#5}%
        }%
    }
\end{tcolorbox}%
}

\newcommand{\ablationsep}{%
  \hspace{1.2pt}%
  {\color{black!30}\vrule width 0.35pt}%
  \hspace{1.2pt}%
}

\definecolor{wacvblue}{rgb}{0.21,0.49,0.74}
\usepackage[pagebackref,breaklinks,colorlinks,allcolors=wacvblue]{hyperref}

\usepackage{xurl}

\def\wacvPaperID{1358} 
\def\confName{WACV}
\def\confYear{2027}

\title{Agentic Multimodal Models for Environmental Hyperspectral Unmixing}

\author{Michał Cholewa$^{*1}$,
Luca Ciampi$^{*2}$, 
Nicola Messina$^2$, 
Przemysław Głomb$^1$, 
Giuseppe Amato$^2$ \\ [0.2em]
$^1$IITiS-PAS, Gliwice, Poland \qquad $^2$ISTI-CNR, Pisa, Italy \qquad $^*$Equal contribution \\
{\tt\small mcholewa@iitis.pl}, {\tt\small luca.ciampi@isti.cnr.it}
}

\begin{document}
\maketitle
\begin{abstract}
Hyperspectral unmixing is a key task in remote sensing that aims to decompose mixed pixels in hyperspectral images into their constituent material signatures, or endmembers, and their fractional abundances. Conventional modular approaches estimate the scene composition through successive model-order estimation, endmember extraction, and abundance estimation stages, whose errors can lead to redundant or ambiguous candidate components and ultimately affect the recovered decomposition. We introduce an algorithm-agnostic, large vision-language model (LVLM)-driven agentic framework that refines the outputs of such pipelines rather than replacing their underlying numerical algorithms. Starting from an initial decomposition, the agent iteratively gathers complementary spectral and spatial evidence through dedicated tools, including spectral-library retrieval and abundance-map visualization, and modifies the active endmember set through merge and discard operations followed by abundance re-estimation. We apply the same refinement procedure to several modular pipelines combining different model-order, extraction, and abundance-estimation methods, and evaluate it on HYDICE Urban, Jasper Ridge, and Stonewall Playa.
Experiments show that the proposed agent consistently improves endmember cardinality and generally improves the recovered spectral signatures and abundance maps across heterogeneous modular pipelines, while remaining competitive with integrated end-to-end unmixing methods, including CNN-AE, uDAS, and R-CoNMF. These results highlight the potential of tool-using LVLM agents to combine spectral and spatial evidence for algorithm-agnostic refinement of physically grounded hyperspectral unmixing decompositions. Code is publicly available at \url{https://anonymous.4open.science/r/agentic-hu/}.
\end{abstract}
    

\section{Introduction}
\label{sec:intro}
Remote sensing enables large-scale observation of natural and human-modified landscapes, supporting applications such as ecological degradation assessment, water-resource management, and land-cover mapping~\cite{Gorelick_2017,Dubovyk_2017,Palmer_2015,Khatami_2016}. Many of these applications require identifying the material composition and biophysical properties of observed surfaces rather than relying only on their visual appearance. However, conventional broadband optical imagery provides limited spectral information for distinguishing materials with similar appearance. Hyperspectral imaging (HSI) addresses this limitation by acquiring dense measurements across hundreds of narrow, contiguous spectral bands for each spatial pixel~\cite{Goetz1985ImagingSpectrometry}. The resulting spectral signatures capture fine-grained absorption and reflectance patterns, enabling detailed material discrimination and quantitative analysis of observed scenes.


Despite their rich spectral information, hyperspectral images are often acquired at spatial resolutions at which individual pixels contain multiple materials~\cite{zhu2017survey}. Consequently, the measured spectrum of a pixel often does not represent a single material, but instead combines contributions from multiple surface components. Hyperspectral unmixing addresses this mixed-pixel problem by decomposing each observed spectrum into constituent spectral signatures, known as \textit{endmembers}, and estimating their corresponding fractional abundances~\cite{BioucasDias2012}. A substantial portion of the hyperspectral unmixing literature performs this decomposition under the assumption that the number of endmembers is known in advance. In operational scenarios, however, the endmember count is generally unavailable and must be inferred from the observed image, along with the endmember signatures and abundance maps. Comparatively fewer works formulate and evaluate unmixing as an end-to-end problem in which all three quantities are recovered without scene-specific reference information. This setting remains challenging because errors, ambiguities, and redundancies in the estimated endmember set directly affect the resulting abundance estimates~\cite{s25082592,DBLP:journals/corr/HahnZ17a,7009875}.

In this work, we present an algorithm-agnostic, large vision-language model (LVLM)-driven agentic framework for refining hyperspectral unmixing decompositions in the challenging setting where the target-scene endmember count, spectral signatures, and abundance maps are unknown \textit{a priori}. The framework refines an initial decomposition produced directly from the hyperspectral image by a modular unmixing pipeline, which may contain redundant or ambiguous candidate endmembers. It can operate with different combinations of model-order estimation, endmember extraction, and abundance estimation methods. Through a ReAct-style tool-using loop~\cite{DBLP:conf/iclr/YaoZYDSN023}, the agent autonomously inspects the current decomposition and selects refinement actions. Specifically, it combines access to an external spectral library with visual inspection of abundance maps overlaid on an RGB composite derived from the hyperspectral image. Based on spectral-library matches and spatial abundance patterns, the agent merges redundant endmembers or discards unsupported candidates. After each update, the abundance maps are re-estimated using the method associated with the underlying pipeline. The resulting system returns the refined endmember set, its cardinality, and the corresponding abundance maps.

We evaluate the proposed framework on three publicly available
hyperspectral unmixing benchmarks representing distinct land-cover
scenarios: the complex urban
landscape of HYDICE Urban~\cite{qian2011hyperspectral,zhu2017survey}, the natural vegetated environment of Jasper Ridge~\cite{zhu2014structured,zhu2017survey}, and the arid, mineral-rich landscape of Stonewall
Playa~\cite{goetz1996understanding}. To assess its algorithm-agnostic applicability, we apply the agent to initial decompositions produced by multiple conventional unmixing pipelines combining different model-order estimation, endmember extraction, and abundance estimation methods. We further compare the resulting systems with existing end-to-end unmixing approaches that start from an initial upper bound on the number of components and progressively identify the active endmember set and estimate the corresponding spectral signatures and abundance maps without access to scene-specific ground truth. Our experiments show that the proposed framework effectively refines initial decompositions produced by different underlying pipelines, improving the estimated endmember sets and corresponding abundance maps. The refined systems are also competitive with existing end-to-end unmixing approaches, outperforming them on several datasets and evaluation metrics. These results highlight the potential of LVLM-driven agents as general-purpose refinement modules for hyperspectral unmixing.

The main contributions of this work are as follows:
\begin{itemize}
    \item We introduce an algorithm-agnostic LVLM-driven agentic framework for hyperspectral unmixing without scene-specific knowledge of the endmember count, spectral signatures, or abundance maps. The framework refines decompositions produced by different conventional pipelines combining model-order estimation, endmember extraction, and abundance estimation methods, ultimately returning the refined endmember set, its cardinality, and the corresponding abundance maps.

    \item We design a ReAct-style tool-using agent~\cite{DBLP:conf/iclr/YaoZYDSN023} that autonomously drives this refinement process. The agent combines retrieval-augmented reasoning over an external spectral library with visual inspection of abundance maps overlaid on an RGB composite of the hyperspectral image. Based on this complementary spectral and spatial evidence, it executes merge and discard operations and triggers abundance re-estimation after each update.

    \item We conduct a systematic evaluation on HYDICE Urban~\cite{qian2011hyperspectral,zhu2017survey}, Jasper Ridge~\cite{zhu2014structured,zhu2017survey}, and Stonewall Playa~\cite{goetz1996understanding}, applying the framework to multiple initial unmixing pipelines and comparing our approach with existing methods under the same unmixing setting. The results show that the proposed framework improves initial decompositions across different underlying pipelines and achieves competitive performance against end-to-end unmixing methods, outperforming them on several datasets and evaluation metrics.
\end{itemize}

\section{Related Work}
\label{sec:related_work}

\subsection{Hyperspectral unmixing for Earth observation}

A standard formulation for hyperspectral unmixing in Earth observation is the linear mixing model, which represents each pixel spectrum as a combination of endmember signatures weighted by their abundances~\cite{BioucasDias2012,Ma_2014}. 
A modular linear unmixing pipeline typically comprises three stages: model-order estimation, endmember extraction, and abundance estimation~\cite{BioucasDias2012}. Representative model-order estimation methods include HySime~\cite{BioucasDiasNascimento2008}, which estimates the signal subspace dimension, and NWHFC~\cite{Chang_2004}, which estimates the virtual dimensionality of hyperspectral data. 
Given this number, representative endmember extraction methods include pure-pixel algorithms such as N-FINDR~\cite{Winter_1999} and VCA~\cite{Nascimento_2005}, as well as minimum-volume methods such as SISAL~\cite{DBLP:conf/whispers/Bioucas-Dias09}. Abundances can then be estimated using FCLSU~\cite{Heinz_2001} or UnDIP~\cite{DBLP:journals/tgrs/RastiKSG22}.
These stage-specific algorithms can be combined into complete modular pipelines~\cite{DBLP:journals/staeors/BernabeSPLBS13,DBLP:journals/staeors/TortiDLP16}. 
Nevertheless, a substantial portion of the unmixing literature estimates endmember signatures and abundance maps while assuming that the number of endmembers is known beforehand~\cite{DBLP:journals/corr/HahnZ17a,s25082592}. The act of providing the reference cardinality bypasses model-order estimation and therefore excludes a key stage of the complete unmixing problem. Although this assumption facilitates the development and evaluation of individual unmixing stages, it may be restrictive in operational settings where the endmember cardinality is not known in advance. 

Integrated unmixing methods address a related end-to-end unmixing setting within a single method-specific formulation rather than by combining separate stage-specific algorithms. These methods initialize an overcomplete model using a prescribed upper bound $K_{\max}$ and progressively identify the active components while estimating their signatures and abundances. They therefore recover all three elements of the decomposition, but require $K_{\max}$ to be specified \textit{a priori}. Specifically, R-CoNMF~\cite{DBLP:journals/tgrs/LiBPL16} suppresses redundant components through collaborative regularization while estimating the mixing matrix and fractional abundances. uDAS~\cite{Qu2019uDAS} employs an untied denoising autoencoder with sparsity regularization to deactivate redundant hidden units while estimating the endmember signatures and abundances. More recently, Alshahrani et al.~\cite{s25082592} combined autoencoder-based unmixing with competitive agglomerative clustering to identify the active endmember set and estimate the corresponding abundance maps.

A separate line of research considers sparse or library-based unmixing, in which each pixel is represented using a small subset of spectra selected from an external dictionary~\cite{5692827,Iordache_2012}. Large spectral libraries can be highly coherent, and their reference signatures may differ from those observed under the acquisition conditions of the target scene, motivating dictionary-pruning and mismatch-aware approaches~\cite{Iordache_2014,Fu_2016,Preston_2025}. Our framework also accesses an external spectral library, but for a different purpose: it is not used as the numerical endmember dictionary for the mixing model. Instead, retrieved reference spectra serve as auxiliary evidence for the LVLM agent and are neither incorporated into the active endmember set nor used to reconstruct the observed image. The decomposition being refined remains derived from the observed hyperspectral image.

\subsection{Large Vision-Language Models and Agentic AI in Remote Sensing}
Recent foundation models align remote sensing imagery with language for captioning, question answering, classification, and spatial grounding, as exemplified by GeoChat~\cite{Kuckreja_2024}. For hyperspectral imagery, SpectralGPT~\cite{DBLP:journals/pami/HongZLLLYYLGJPGBC24} and HyperSIGMA~\cite{DBLP:journals/pami/WangHJMYXQMSLFCHYZXLZWD25} learn transferable spatial--spectral representations, while HyperCap~\cite{Das_2026} and HM-Bench~\cite{DBLP:journals/corr/abs-2604-08884} respectively introduce textual annotations and evaluate multimodal spatial--spectral reasoning. Since general-purpose multimodal models cannot directly process raw hyperspectral cubes, these approaches rely on suitable visual or textual representations of the spectral data. Nevertheless, they focus on hyperspectral understanding rather than end-to-end unmixing.

Agentic AI extends language and vision-language models with iterative reasoning and external tool use, as formalized by ReAct~\cite{DBLP:conf/iclr/YaoZYDSN023}. In remote sensing, RS-Agent~\cite{DBLP:journals/corr/abs-2406-07089} combines an LLM controller, retrieval, and specialized image-processing tools, while GeoLLM-QA~\cite{DBLP:journals/corr/abs-2405-00709} and ThinkGeo~\cite{DBLP:journals/corr/abs-2505-23752} evaluate multi-step, tool-based geospatial reasoning. These systems primarily address conventional image-understanding and geospatial-analysis tasks. The use of tool-using LVLM agents for hyperspectral unmixing therefore remains largely unexplored. Our framework employs an LVLM as an algorithm-agnostic controller that iteratively refines decompositions produced by modular unmixing pipelines. The agent does not replace the numerical algorithms of the underlying unmixing pipeline; instead, it retrieves reference spectra from an external library, inspects abundance maps overlaid on an RGB composite derived from the hyperspectral image, updates the candidate endmember set through merge and discard operations, and triggers abundance re-estimation after each refinement. In this way, the proposed framework combines spectral and spatial evidence to refine initial decompositions generated by different modular unmixing pipelines.

\begin{figure*}[t]
    \centering
        \includegraphics[width=0.8\textwidth]{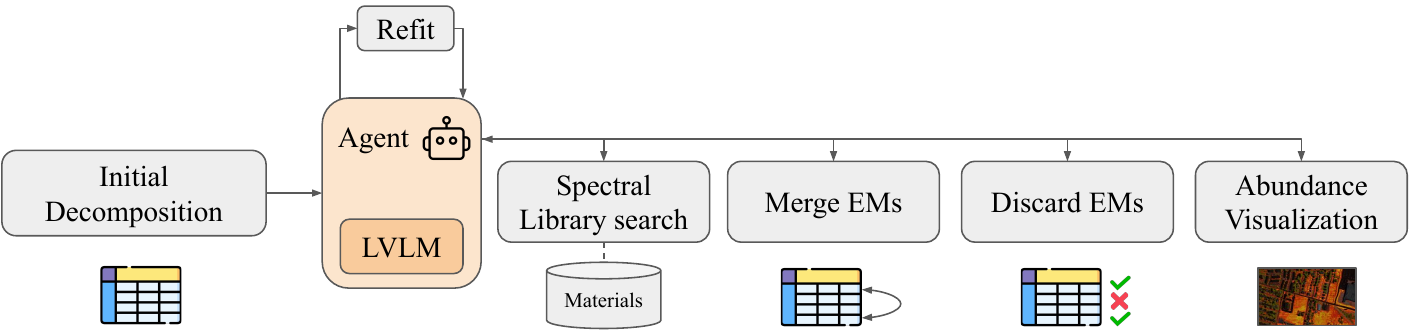}
        \caption{\textbf{Overview of the proposed LVLM-driven agentic refinement framework.} Starting from an initial decomposition $(K_0,\mathbf{M}_0,\mathbf{A}_0)$, the LVLM agent iteratively interacts with dedicated tools to gather complementary evidence and refine the candidate endmembers. Spectral-library search provides external material references, while abundance-map visualization supports their spatial inspection. Based on this evidence, the agent can merge redundant endmembers or discard unsupported candidates. Each state-changing action updates the active endmember set and is followed by abundance re-estimation (\textit{Refit}) using the method associated with the underlying modular pipeline. The resulting decomposition becomes the current state for the next reasoning step.} 
        \label{fig:pipeline}
\end{figure*}

\section{Methodology}
\label{sec:method}

This section presents the proposed algorithm-agnostic, LVLM-driven agentic framework for refining hyperspectral unmixing decompositions. We consider the setting in which no scene-specific reference decomposition is provided: the true endmember count, spectral signatures, and abundance maps are all unknown \textit{a priori}. Rather than replacing the numerical unmixing algorithms, the proposed framework operates on an initial decomposition $(K_0,\mathbf{M}_0,\mathbf{A}_0)$ produced directly from the hyperspectral image by a modular pipeline. To assess its independence from the underlying numerical methods, we apply the same refinement process to initial decompositions generated by several representative combinations of well-established model-order estimation, endmember extraction, and abundance estimation algorithms. 

\subsection{Notation and Linear Mixing Model}

Let $\mathbf{Y} \in \mathbb{R}^{L \times N}$ denote the hyperspectral data matrix, where $L$ is the number of spectral bands and $N$ is the number of pixels obtained after reshaping the spatial dimensions of the hyperspectral cube. Each column $\mathbf{y}_i \in \mathbb{R}^{L}$ represents the observed spectrum of the $i$-th pixel.
Under the Linear Mixing Model (LMM), each observed spectrum is represented as a linear combination of $K$ pure material spectra, referred to as endmembers:
\begin{equation}
\mathbf{Y} = \mathbf{M}\mathbf{A} + \mathbf{E},
\end{equation}
where $\mathbf{M} \in \mathbb{R}^{L \times K}$ is the endmember matrix containing the spectral signatures, $\mathbf{A} \in \mathbb{R}^{K \times N}$ is the abundance matrix containing the per-pixel proportions, and $\mathbf{E} \in \mathbb{R}^{L \times N}$ accounts for additive noise and modeling errors.

To retain physical meaning, the abundance coefficients satisfy the Abundance Non-negativity Constraint (ANC):
\begin{equation}
a_{ki} \geq 0, \qquad \forall k \in \{1,\ldots,K\}, \quad \forall i \in \{1,\ldots,N\}.
\end{equation}
When each observed pixel is assumed to be fully explained by the selected endmembers, the coefficients also satisfy the Abundance Sum-to-one Constraint (ASC):
\begin{equation}
\sum_{k=1}^{K} a_{ki} = 1, \qquad \forall i \in \{1,\ldots,N\}.
\end{equation}

\subsection{Initial Modular Unmixing Pipelines}
\label{ref:initial-decomposition}

We obtain the initial decompositions $(K_0,\mathbf{M}_0,\mathbf{A}_0)$ by combining representative model-order estimation, endmember extraction, and abundance estimation methods into complete modular pipelines, as detailed below.

\paragraph{Model-order estimation.}
We consider HySime~\cite{BioucasDiasNascimento2008} and NWHFC~\cite{Chang_2004} to estimate the initial number of candidate endmembers $K_0$ directly from the observed hyperspectral image. HySime models each observed spectrum as $\mathbf{y}_i=\mathbf{x}_i+\mathbf{n}_i$, where $\mathbf{x}_i$ denotes the noise-free signal component and $\mathbf{n}_i$ additive noise. From estimates of the data and noise correlation matrices, $\mathbf{R}_y$ and $\mathbf{R}_n$, it obtains $\mathbf{R}_x = \mathbf{R}_y-\mathbf{R}_n$, and selects the signal subspace that minimizes the expected projection error. Its dimension defines the corresponding estimate $K_0$. NWHFC follows a different criterion based on virtual dimensionality: after whitening the observations using the estimated noise statistics, it tests for spectrally distinct components, using the resulting dimensionality as $K_0$. 

\paragraph{Endmember extraction.}
Given the estimated cardinality $K_0$, we employ either VCA~\cite{Nascimento_2005} or SISAL~\cite{DBLP:conf/whispers/Bioucas-Dias09} to obtain the initial endmember matrix $\mathbf{M}_0\in\mathbb{R}^{L\times K_0}$. VCA is a geometric method based on the pure-pixel assumption and identifies candidate endmembers among extreme observations in the projected spectral space. SISAL instead follows a minimum-volume formulation and estimates a simplex whose vertices represent the endmembers, using soft constraints to accommodate noise and outliers. 

\paragraph{Abundance estimation.}
Given $\mathbf{M}_0$, we estimate the corresponding abundance matrix $\mathbf{A}_0\in\mathbb{R}^{K_0\times N}$ using either FCLSU~\cite{Heinz_2001} or UnDIP~\cite{DBLP:journals/tgrs/RastiKSG22}. FCLSU solves
\begin{equation}
\mathbf{A}_0 =
\arg\min_{\mathbf{A}}
\left\|\mathbf{Y}-\mathbf{M}_0\mathbf{A}\right\|_F^2
\end{equation}
subject to the ANC and ASC constraints. UnDIP instead estimates the abundance maps through a convolutional deep image prior, exploiting their spatial structure while keeping the extracted endmember matrix fixed. 

\paragraph{Pipeline configurations.}
We consider all eight modular pipelines obtained by combining the two model-order estimators, two endmember-extraction methods, and two abundance-estimation methods. These configurations are not proposed as new unmixing methods; rather, they provide heterogeneous initial decompositions for evaluating the algorithm-agnostic applicability of the same refinement framework.

\subsection{LVLM-driven Agentic Refinement}
\label{sec:react-agent-pipeline}

Starting from an initial decomposition $(K_0,\mathbf{M}_0,\mathbf{A}_0)$ generated by one of the modular pipelines described in Section~\ref{ref:initial-decomposition}, the proposed LVLM agent iteratively refines the active endmember set through a ReAct-style tool-using loop~\cite{DBLP:conf/iclr/YaoZYDSN023}, as shown in Figure~\ref{fig:pipeline}. At iteration $t$, the current decomposition is denoted by 

\begin{equation}
(K^{(t)},\mathbf{M}^{(t)},\mathbf{A}^{(t)}),
\end{equation}

\noindent with $(K^{(0)},\mathbf{M}^{(0)},\mathbf{A}^{(0)})=(K_0,\mathbf{M}_0,\mathbf{A}_0)$. Rather than directly processing the hyperspectral vectors in its language context, the LVLM interacts with the numerical decomposition through specialized tools that provide spectral and spatial evidence or modify the active endmember set. Tool outputs are returned to the model as textual or visual observations and become part of the context used to select subsequent actions.

The agent maintains a structured state containing the currently active endmembers, their identifiers, candidate material labels obtained during spectral retrieval, and the information required to track refinement operations. 
At each reasoning step, the agent autonomously decides which candidate endmembers require further inspection, which tools should be invoked, and whether the current decomposition should be modified. The available tools comprise \nolinkurl{library\_search}, \nolinkurl{compute\_abundance}, \nolinkurl{merge\_endmembers}, and \nolinkurl{discard\_endmember}.

\paragraph{Spectral-library retrieval.}
The \nolinkurl{library\_search} tool provides external spectral evidence for a selected active endmember. Given its current signature $\mathbf{m}^{(t)}_k$, the tool aligns the spectrum with the wavelength grid of an external spectral library and ranks the reference signatures according to the Spectral Angle Distance (SAD),
\begin{equation}
\operatorname{SAD}(\mathbf{m},\mathbf{r}) =
\arccos\left(
\frac{\mathbf{m}^{\top}\mathbf{r}}
{\|\mathbf{m}\|_2\|\mathbf{r}\|_2}
\right),
\end{equation}
where $\mathbf{r}$ denotes a reference spectrum from the library. The retrieved matches, together with their material categories and spectral distances, are returned to the LVLM and can be associated with the active endmember as candidate semantic labels. This provides the agent with retrieval-augmented spectral knowledge while preserving the data-driven nature of the unmixing decomposition: retrieved library spectra are used only as auxiliary semantic evidence and are neither inserted into $\mathbf{M}^{(t)}$ nor directly used to reconstruct $\mathbf{Y}$.

\paragraph{Spatial abundance inspection.}
The \nolinkurl{compute\_abundance} tool provides complementary spatial evidence about the active endmembers. For a selected component, the corresponding row of $\mathbf{A}^{(t)}$ is reshaped according to the spatial dimensions of the hyperspectral cube and rendered as a heatmap overlaid on an RGB composite derived from the same hyperspectral image. The resulting visualization is returned to the LVLM, allowing it to inspect the spatial support and distribution of the candidate material in the context of the observed scene. In particular, the agent can compare abundance patterns across different candidates and jointly reason about spectral-library compatibility and spatial evidence when determining whether two candidates are redundant or whether a component is insufficiently supported.

\paragraph{Agentic Endmember refinement.}
The tools for pruning the endmembers state table are \nolinkurl{merge\_endmembers} and \nolinkurl{discard\_endmember}. A merge operation combines active candidates judged to represent redundant components and replaces them with an abundance-weighted average of their signatures, assigning greater weight to components with higher total abundance mass. Alternative merge strategies are evaluated in the supplementary material. A discard operation instead removes a selected candidate from the active set. In either case, the operation produces an updated endmember matrix $\mathbf{M}^{(t+1)}$ and a corresponding reduced cardinality $K^{(t+1)}$. The abundance maps are then re-estimated using the abundance-estimation method associated with the underlying modular pipeline:
\begin{equation}
\mathbf{A}^{(t+1)} =
\mathcal{U}_{\pi}\!\left(\mathbf{Y},\mathbf{M}^{(t+1)}\right),
\end{equation}
where $\mathcal{U}_{\pi}$ denotes the abundance estimator employed by pipeline $\pi$. Consequently, every accepted merge or discard modifies the numerical decomposition itself, and all subsequent spectral and spatial observations are obtained from the updated state $(K^{(t+1)},\mathbf{M}^{(t+1)},\mathbf{A}^{(t+1)})$.

The refinement proceeds iteratively as the LVLM alternates between gathering evidence and executing state-changing operations. Unlike a procedure based on a fixed sequence of diagnostic tests or predefined merge and discard proposals, the agent determines which endmembers to inspect, which spectral or spatial evidence to acquire, and whether a refinement action is warranted based on the observations accumulated during the interaction. After every state-changing operation, the updated decomposition is exposed to subsequent tool calls, allowing the agent to reconsider previous evidence in light of the new active endmember set. When the agent determines that no further refinement is required, the current state is returned as the final decomposition $(\widehat{K},\widehat{\mathbf{M}},\widehat{\mathbf{A}})$.

The current action space makes the refinement one-sided: merge and discard operations can consolidate or remove components represented in the initial decomposition, but they cannot add new independent components or increase the endmember cardinality. Consequently, $\widehat{K}\leq K_0$, and the quality of the final decomposition remains partly dependent on the coverage provided by the initial candidate set. We evaluate the same refinement process under heterogeneous initializations to characterize this dependence and assess its applicability across different backbone models. 

\section{Experimental Setting}

\paragraph{Setup and Implementation Details.}
For each experiment, an initial decomposition $(K_0,\mathbf{M}_0,\mathbf{A}_0)$ is first generated from the hyperspectral image by one of the modular unmixing pipelines described in Section~\ref{ref:initial-decomposition}, and the same LVLM-driven refinement framework is then applied to it. The LVLM employed by the agent is Qwen-3.6 27B, with its native reasoning mode disabled, while additional models are considered in the ablation study of Section~\ref{sec:ablation}. The model temperature is set to $0$ to improve reproducibility. The \nolinkurl{library\_search} tool queries the \textit{USGS Spectral Library}~\cite{kokaly2017usgs}. For spatial inspection, each abundance map is combined with the corresponding RGB composite using a 50/50 alpha blend, while the RGB composite is also provided independently to the agent as visual context.

Because the initial decompositions may depend on stochastic components of the underlying pipelines, each experiment is repeated over 10 runs. Within each run, the agent is applied to exactly the same initial decomposition used for the corresponding unrefined baseline, yielding a paired comparison between the initial and refined solutions. Results are reported as the mean across runs.

\begin{figure}[t]
    \centering
    \setlength{\abovecaptionskip}{4pt}

    \makebox[\columnwidth][c]{%
    \scalebox{0.94}{%
    \begin{minipage}{\columnwidth}
        \centering

        \datasetrow
            {HYDICE Urban}
            {UrbanLine}
            {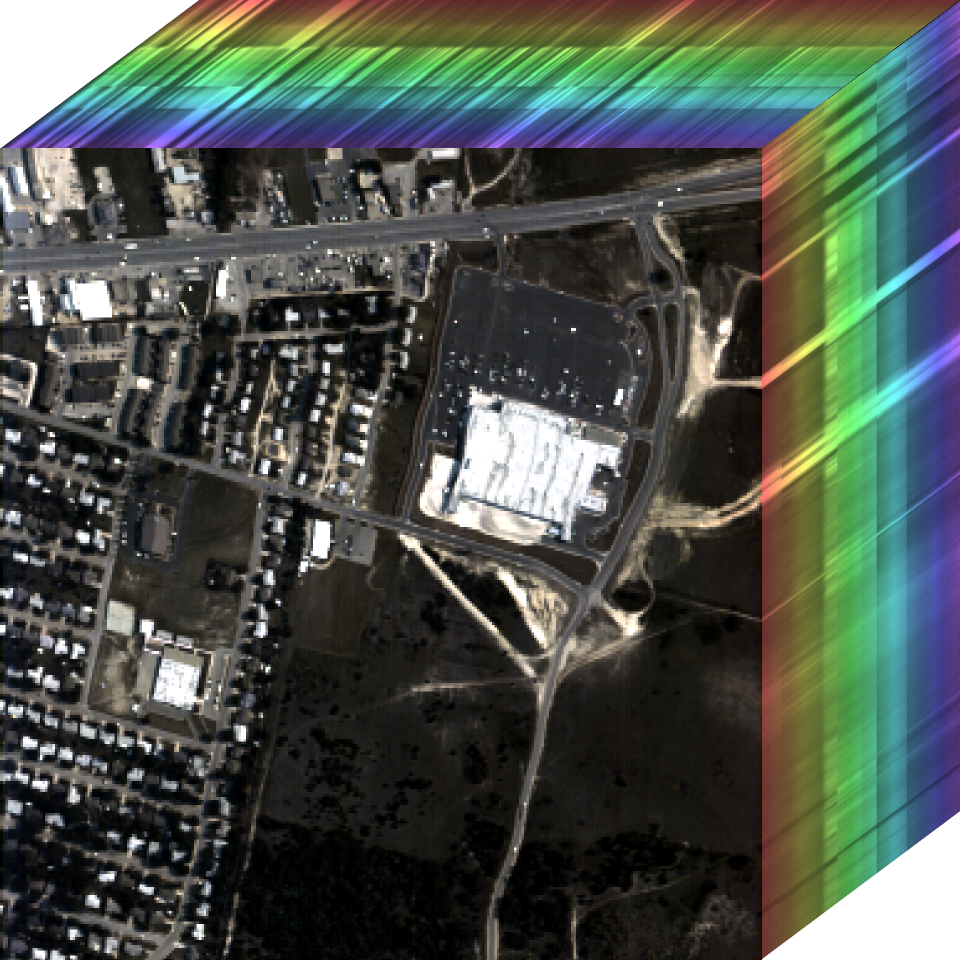}
            {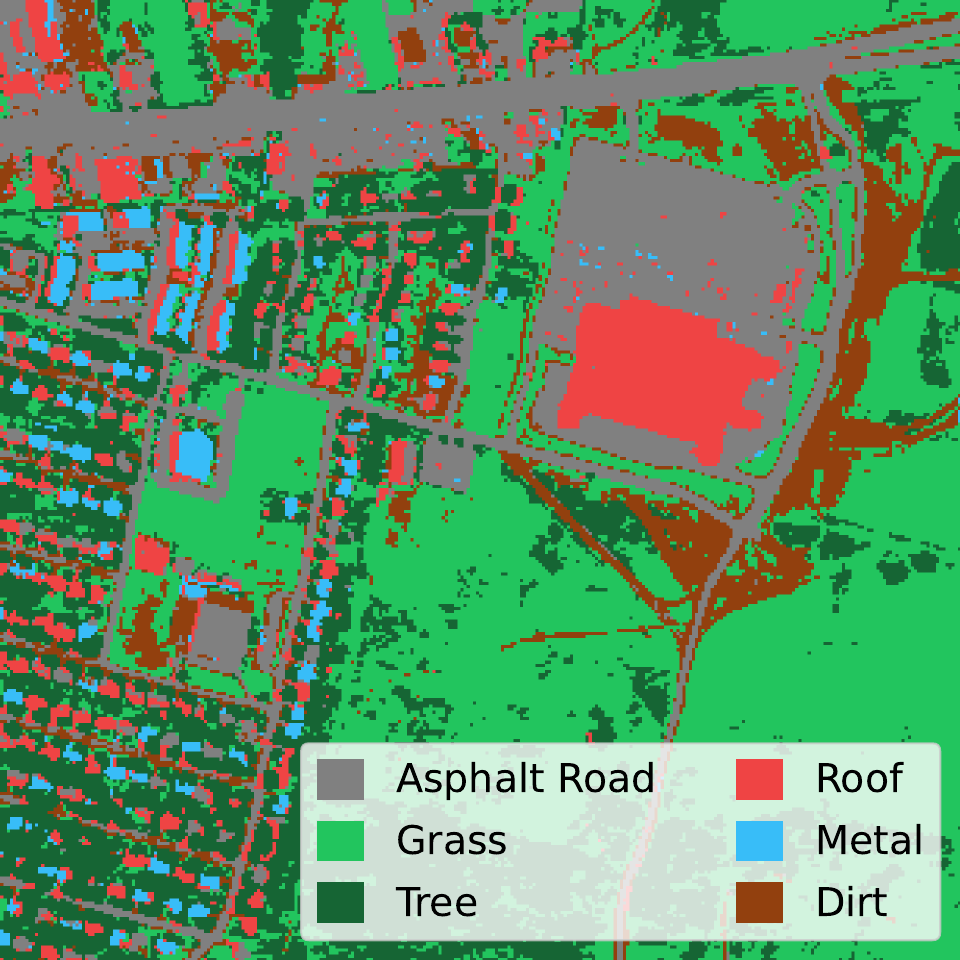}
            {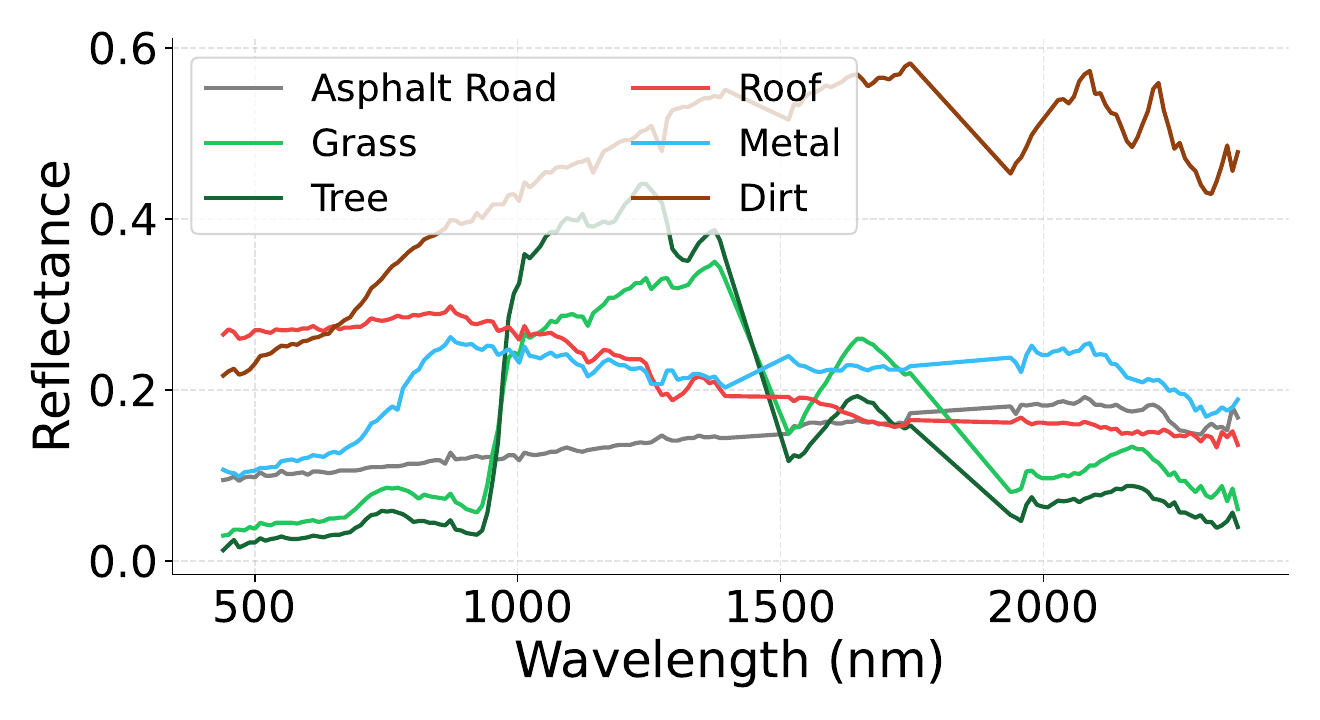}

        \datasetrow
            {Jasper Ridge}
            {JasperLine}
            {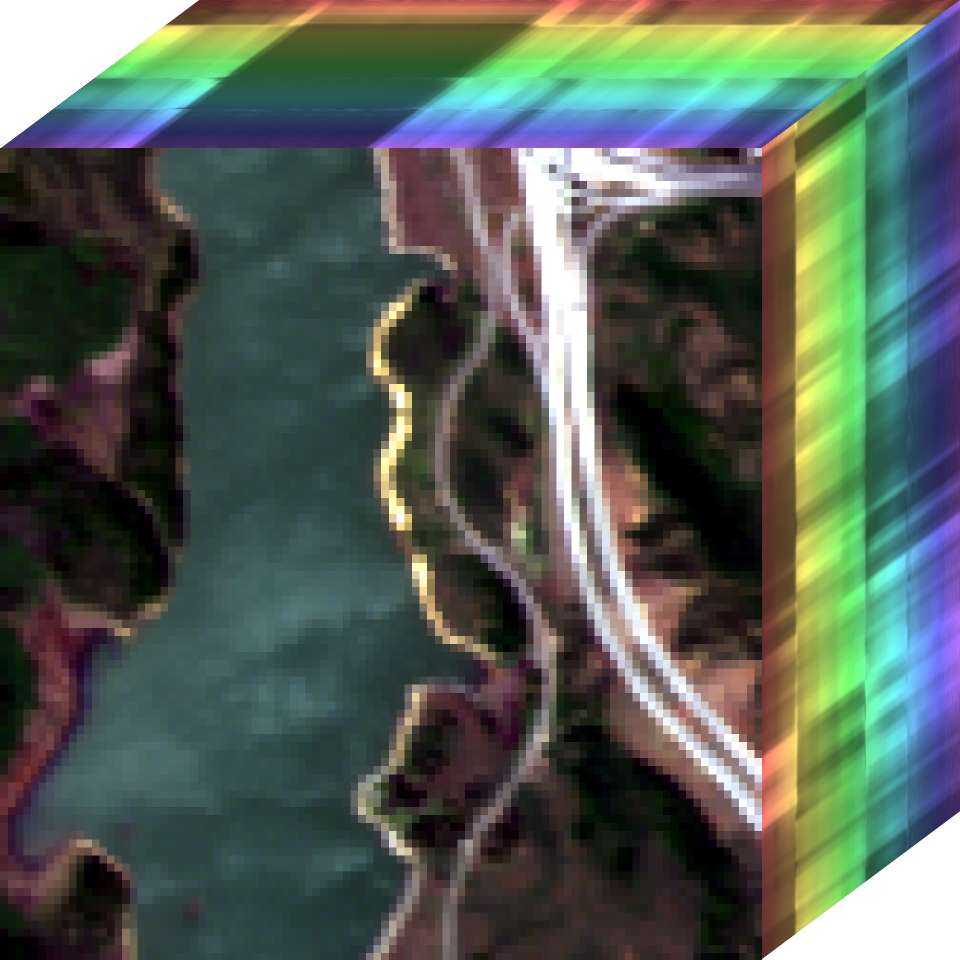}
            {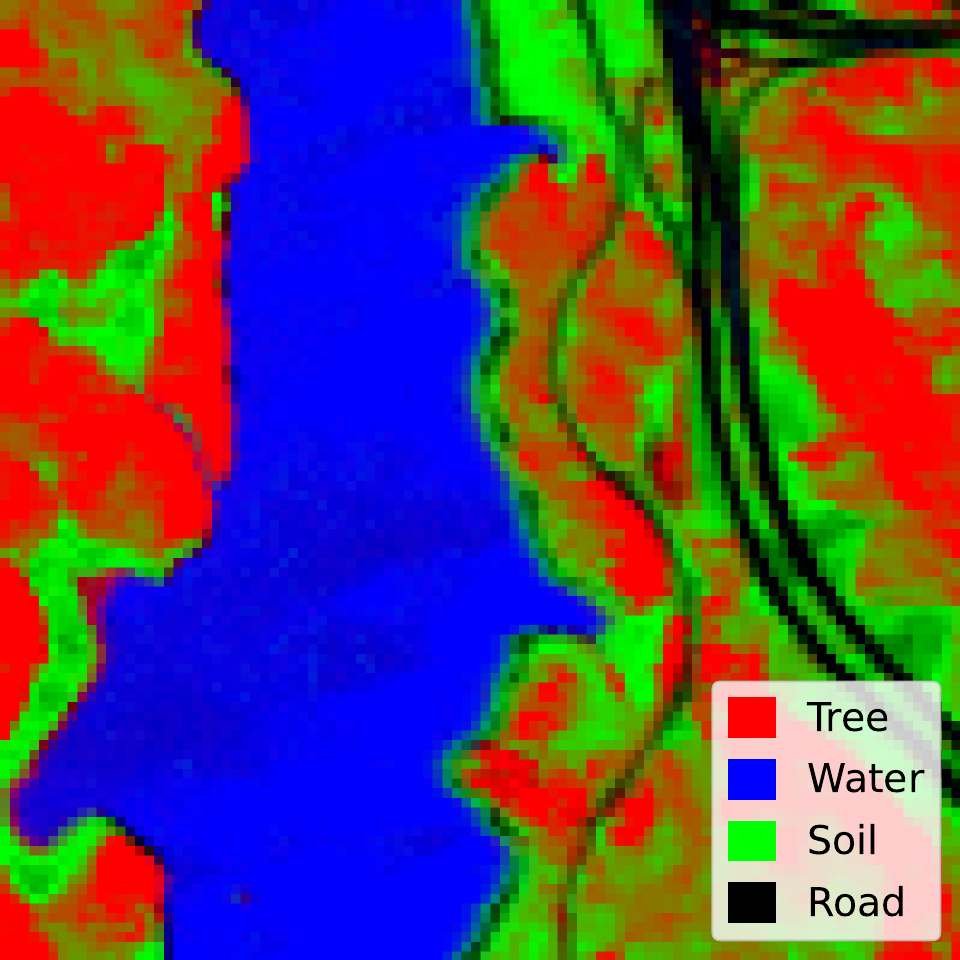}
            {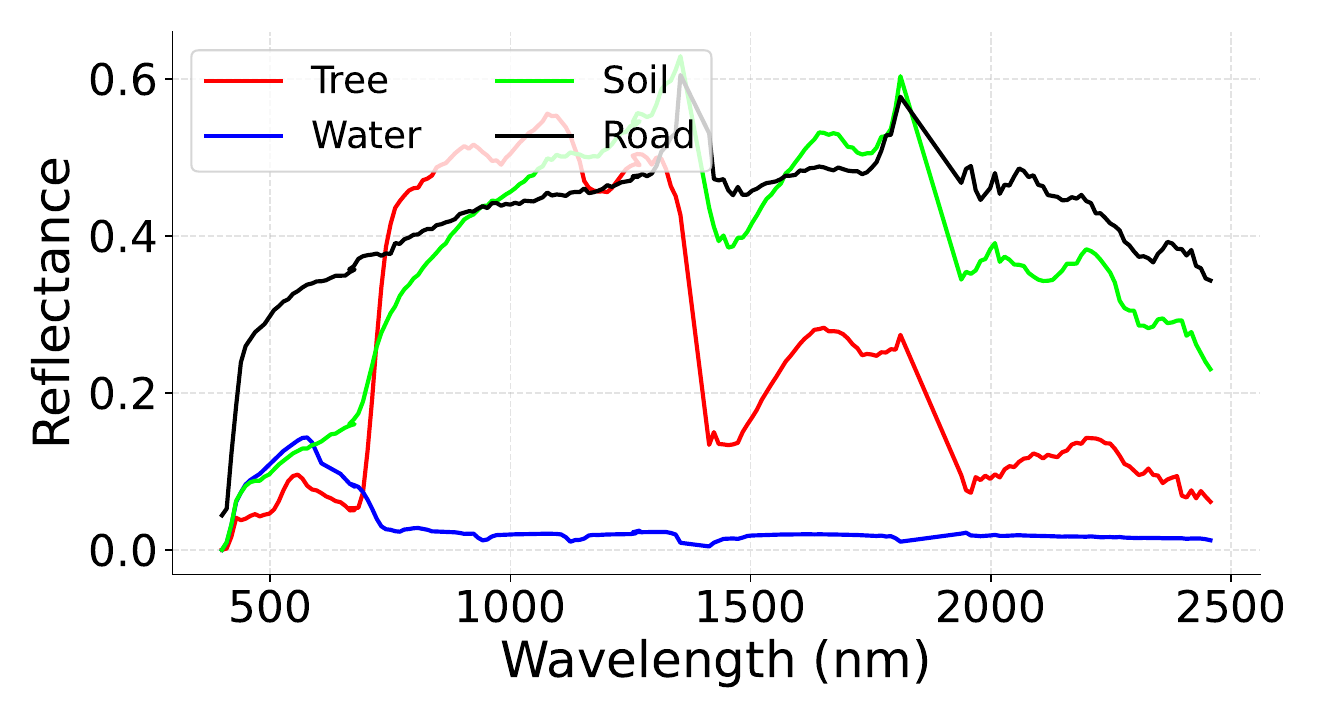}

        \datasetrow
            {Stonewall Playa}
            {StonewallLine}
            {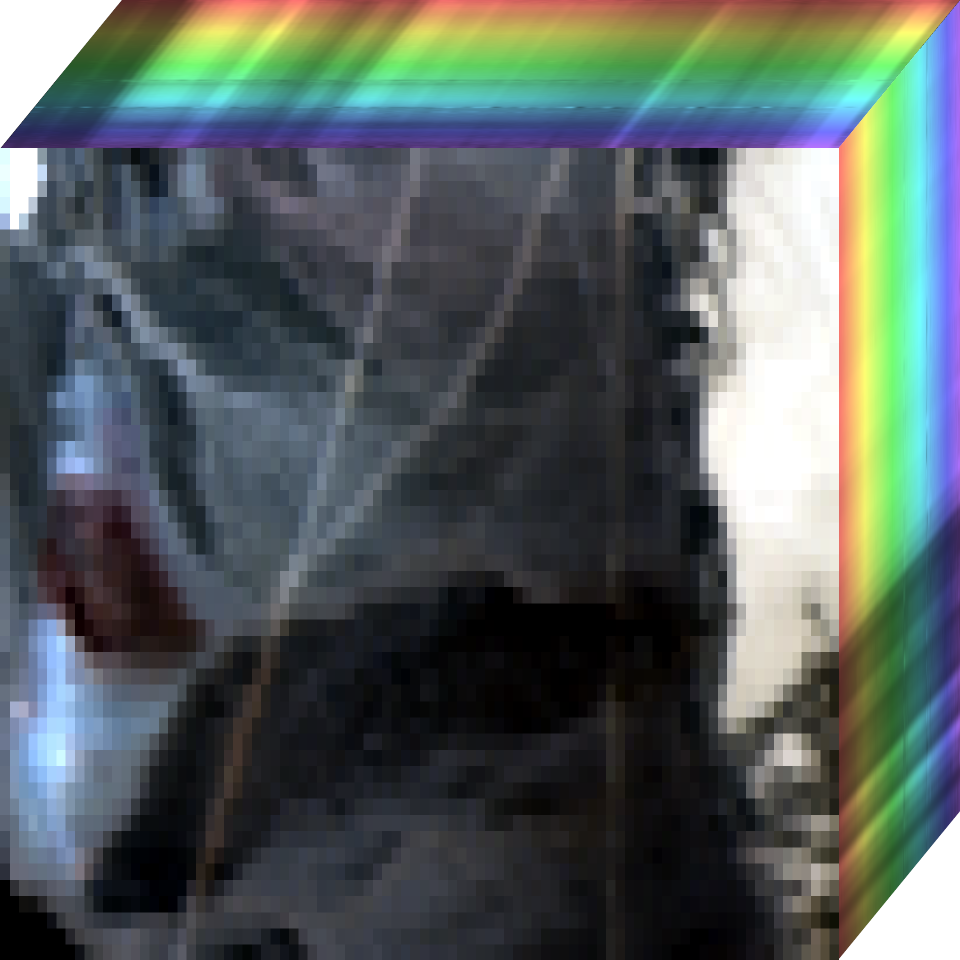}
            {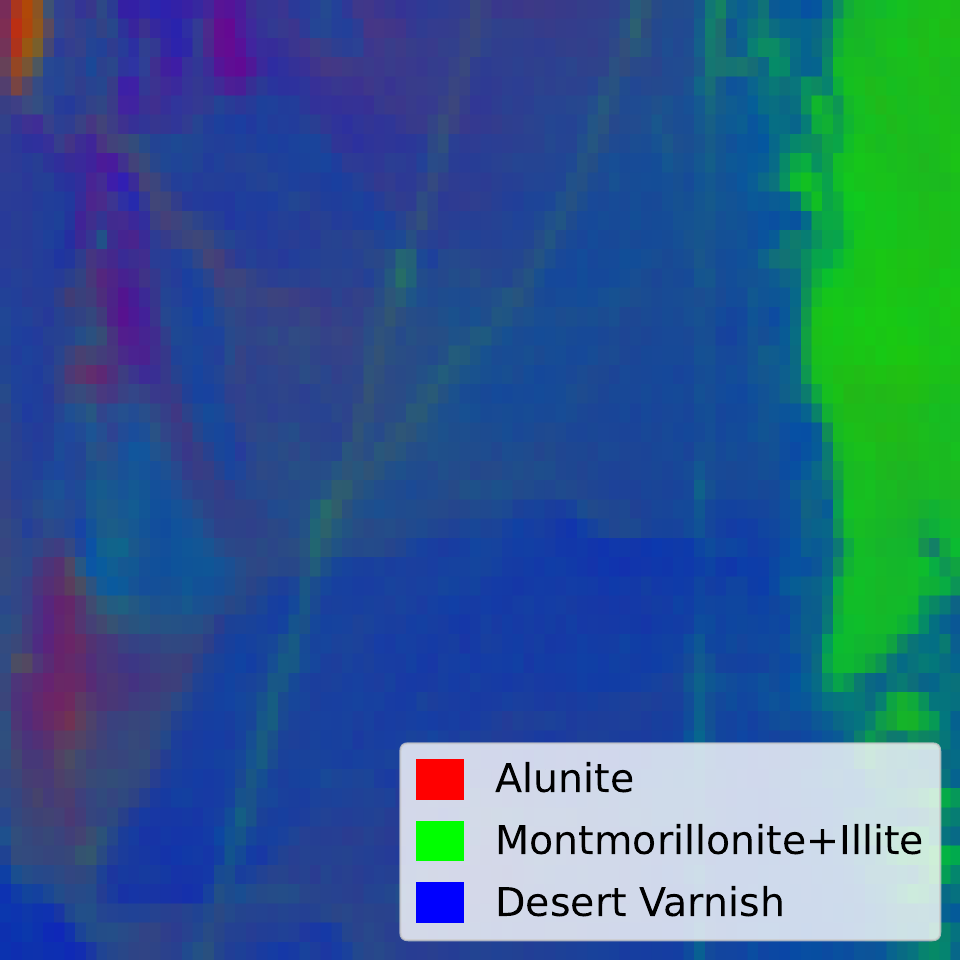}
            {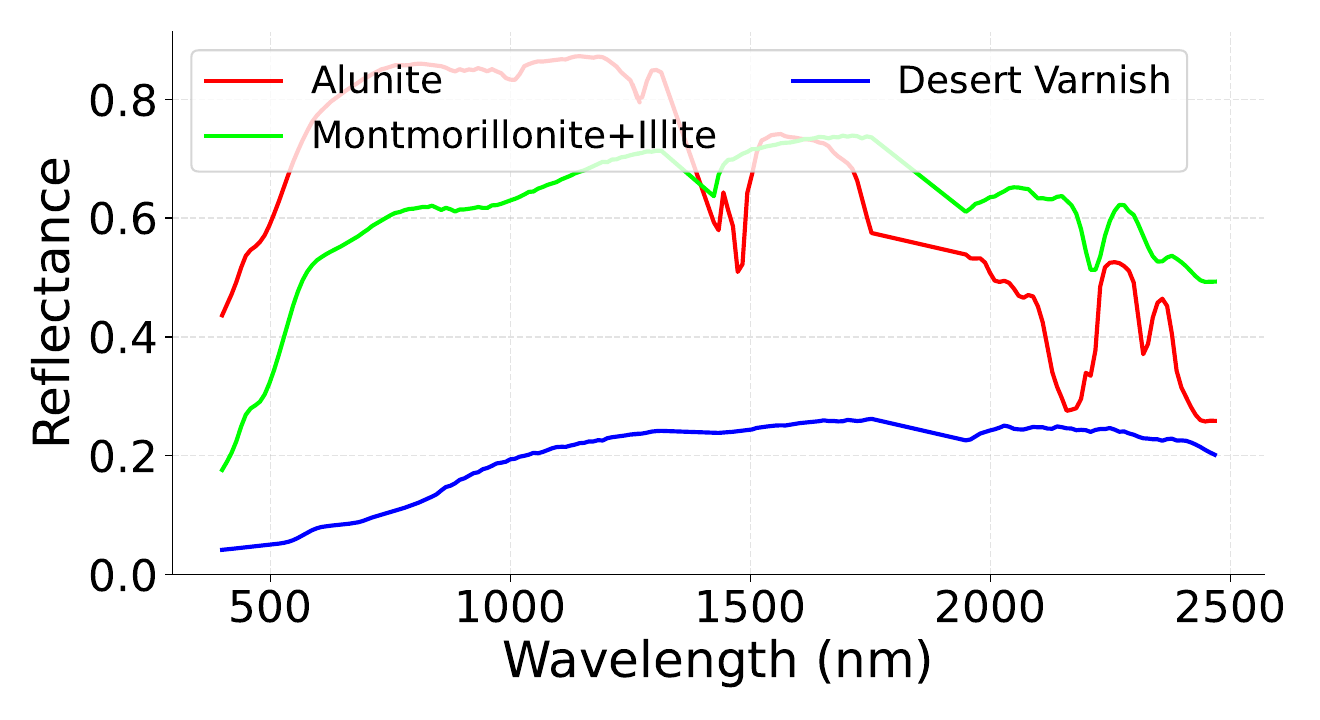}

    \end{minipage}%
    }}

    \caption{\textbf{Overview of the HYDICE Urban, Jasper Ridge, and Stonewall
    Playa datasets.} Each row shows false-color images, abundance maps,
    and reference spectral signatures.
    }

    \label{fig:datasets_overview}
\end{figure}

\paragraph{Datasets.}
We evaluate the framework on three standard hyperspectral unmixing benchmarks representing different land-cover scenarios (Fig.~\ref{fig:datasets_overview}); results on an additional dataset---Samson~\cite{zhu2017survey}---are in the supplementary material. All use approximately 400--2500~nm wavelength range.

\noindent\textbf{HYDICE Urban~\cite{qian2011hyperspectral,zhu2017survey}.}
A $307\times307$ scene of Copperas Cove, Texas, with 210 bands (162 retained) and $2$~m GSD. We use the standard 6-endmember reference (asphalt, grass, trees, roofs, dirt, and metal).

\noindent\textbf{Jasper Ridge~\cite{zhu2014structured,zhu2017survey}.}
An AVIRIS scene with 224 bands (198 retained) and 20~m GSD. We use the standard $100\times100$ ROI with 4 endmembers: road, soil, water, and trees.

\noindent\textbf{Stonewall Playa~\cite{goetz1996understanding}.}
An AVIRIS scene from the Cuprite mining district, Nevada. We use a $50\times90$ subset with 187 bands and 3 reference signatures: Alunite, Montmorillonite/Illite, and desert varnish.

\paragraph{Evaluation Metrics.}
We assess complementary aspects of the unmixing task using four widely
adopted measures (formal definitions and implementation details are provided in the supplementary material): (i) \emph{Cardinality error ($\mathbf{\Delta K}$)}: measures the absolute difference between the estimated and reference numbers of endmembers; (ii) \emph{Mean spectral angle distance (\textbf{mSAD})}: measures the angular discrepancy between estimated and reference endmembers~\cite{Keshava2002}; (iii) 
\emph{Abundance RMSE (\textbf{aRMSE})}: measures the discrepancy between estimated and reference abundance maps~\cite{Paura2023Benchmark}; (iv) \emph{Reconstruction RMSE (\textbf{rRMSE})}: measures how accurately the decomposition reconstructs the hyperspectral image~\cite{Paura2023Benchmark}.

\begin{table*}[t]
\captionsetup{skip=4pt}
\centering
\small
\renewcommand{\arraystretch}{1.05}

\caption{\textbf{Initial modular decompositions and their refinement by the LVLM-driven agent.} For each modular pipeline, \emph{Initial} carries its original performance, while \emph{+Agent} carries the results obtained after agentic refinement. Best values in \textbf{bold}.}

\label{tab:initial-agent-results}

\resizebox{\linewidth}{!}{%
\begin{tabular}{
@{}
l
!{\datasetsep}
l
!{\datasetsep}
>{\columncolor{UrbanBG}}c
>{\columncolor{UrbanBG}}c
>{\columncolor{UrbanBG}}c
>{\columncolor{UrbanBG}}c
!{\datasetsep}
>{\columncolor{JasperBG}}c
>{\columncolor{JasperBG}}c
>{\columncolor{JasperBG}}c
>{\columncolor{JasperBG}}c
!{\datasetsep}
>{\columncolor{StonewallBG}}c
>{\columncolor{StonewallBG}}c
>{\columncolor{StonewallBG}}c
>{\columncolor{StonewallBG}}c
@{}
}
\toprule

& &
\multicolumn{4}{>{\columncolor{UrbanBG}}c}{\textbf{HYDICE Urban}} &
\multicolumn{4}{>{\columncolor{JasperBG}}c}{\textbf{Jasper Ridge}} &
\multicolumn{4}{>{\columncolor{StonewallBG}}c}{\textbf{Stonewall Playa}} \\

\cmidrule(lr){3-6}
\cmidrule(lr){7-10}
\cmidrule(l){11-14}

\textbf{Pipeline}
& \textbf{Stage}
& $\Delta K\,\downarrow$
& mSAD $\downarrow$
& aRMSE $\downarrow$
& rRMSE $\downarrow$
& $\Delta K\,\downarrow$
& mSAD $\downarrow$
& aRMSE $\downarrow$
& rRMSE $\downarrow$
& $\Delta K\,\downarrow$
& mSAD $\downarrow$
& aRMSE $\downarrow$
& rRMSE $\downarrow$ \\

\midrule

\multirow{2}{*}{HySime--VCA--FCLSU}
& Initial
& 21.0 & 0.466 & 0.373 & 0.123
& 13.0 & \textbf{0.513} & 0.505 & 0.078
& 9.0 & 0.204 & 0.440 & 0.166 \\

& + Agent
& \textbf{6.6} & \textbf{0.448} & \textbf{0.363} & \textbf{0.097}
& \textbf{3.1} & 0.524 & \textbf{0.503} & \textbf{0.049}
& \textbf{2.3} & \textbf{0.199} & \textbf{0.378} & \textbf{0.064} \\

\addlinespace[2pt]

\multirow{2}{*}{HySime--VCA--UnDIP}
& Initial
& 21.0 & 0.466 & 0.387 & 0.123
& 13.0 & 0.513 & \textbf{0.498} & \textbf{0.046}
& 9.0 & 0.204 & 0.458 & 0.059 \\

& + Agent
& \textbf{2.1} & \textbf{0.441} & \textbf{0.376} & \textbf{0.077}
& \textbf{2.5} & \textbf{0.502} & 0.505 & 0.049
& \textbf{1.1} & \textbf{0.197} & \textbf{0.291} & \textbf{0.033} \\

\addlinespace[2pt]

\multirow{2}{*}{HySime--SISAL--FCLSU}
& Initial
& 21.0 & 0.775 & \textbf{0.322} & 0.118 
& 13.0 & 0.863 & \textbf{0.429} & 0.172
& 9.0 & 0.463 & 0.313 & 0.074 \\

& + Agent
& \textbf{2.0} & \textbf{0.507} & 0.352 & \textbf{0.097}
& \textbf{3.3} & \textbf{0.663} & 0.441 & \textbf{0.105}
& \textbf{1.2} & \textbf{0.278} & \textbf{0.301} & \textbf{0.067} \\

\addlinespace[2pt]

\multirow{2}{*}{HySime--SISAL--UnDIP}
& Initial
& 21.0 & 0.814 & 0.404 & 0.321
& 13.0 & 0.863 & \textbf{0.445} & 0.145
& 9.0 & 0.463 & \textbf{0.323} & 0.075 \\

& + Agent
& \textbf{3.4} & \textbf{0.731} & \textbf{0.371} & \textbf{0.289}
& \textbf{2.2} & \textbf{0.699} & 0.481 & \textbf{0.098}
& \textbf{1.1} & \textbf{0.428} & 0.348 & \textbf{0.069} \\

\addlinespace[2pt]

\multirow{2}{*}{NWHFC--VCA--FCLSU}
& Initial
& 31.0 & 0.455 & 0.383 & 0.138
& 9.0 & 0.515 & 0.489 & 0.063
& 5.0 & 0.207 & 0.367 & 0.076 \\

& + Agent
& \textbf{6.2} & \textbf{0.451} & \textbf{0.376} & \textbf{0.090}
& \textbf{1.3} & \textbf{0.508} & \textbf{0.484} & \textbf{0.050}
& \textbf{2.0} & \textbf{0.199} & \textbf{0.332} & \textbf{0.068} \\

\addlinespace[2pt]

\multirow{2}{*}{NWHFC--VCA--UnDIP}
& Initial
& 31.0 & \textbf{0.455} & \textbf{0.378} & 0.123
& 9.0 & 0.515 & \textbf{0.498} & 0.058
& 5.0 & 0.207 & 0.434 & 0.072 \\

& + Agent
& \textbf{9.7} & 0.482 & 0.382 & \textbf{0.089}
& \textbf{1.3} & \textbf{0.510} & 0.499 & \textbf{0.042}
& \textbf{2.2} & \textbf{0.202} & \textbf{0.396} & \textbf{0.045} \\

\addlinespace[2pt]

\multirow{2}{*}{NWHFC--SISAL--FCLSU}
& Initial
& 31.0 & 0.683 & 0.336 & 0.177
& 9.0 & 0.889 & 0.412 & 0.135
& 5.0 & 0.813 & 0.319 & 0.304 \\

& + Agent
& \textbf{8.1} & \textbf{0.590} & \textbf{0.330} & \textbf{0.105}
& \textbf{3.3} & \textbf{0.854} & \textbf{0.399} & \textbf{0.122}
& \textbf{0.9} & \textbf{0.551} & \textbf{0.231} & \textbf{0.185} \\

\addlinespace[2pt]

\multirow{2}{*}{NWHFC--SISAL--UnDIP}
& Initial
& 31.0 & 0.683 & \textbf{0.335} & 0.158
& 9.0 & 0.889 & 0.449 & 0.115
& 5.0 & 0.813 & 0.301 & 0.275 \\

& + Agent
& \textbf{11.0} & \textbf{0.608} & 0.352 & \textbf{0.086}
& \textbf{2.5} & \textbf{0.806} & \textbf{0.421} & \textbf{0.099}
& \textbf{0.8} & \textbf{0.569} & \textbf{0.273} & \textbf{0.149} \\

\bottomrule
\end{tabular}%
}
\end{table*}

\begin{table*}[t]
\captionsetup{skip=4pt}
\centering
\small
\renewcommand{\arraystretch}{1.08}

\caption{\textbf{Comparison under HySime and NWHFC initialization.}
Agent-refined modular pipelines and state-of-the-art end-to-end unmixing methods
(CNN-AE \cite{Palsson2021CNNAEU}, uDAS \cite{Qu2019uDAS}, and
R-CoNMF \cite{DBLP:journals/tgrs/LiBPL16}) are evaluated using initial
cardinalities estimated by HySime and NWHFC. Best and second-best values are \textbf{bolded} and
\underline{underlined}, respectively. \emph{Best} counts the metric-wise
wins.}

\label{tab:sota}

\resizebox{\linewidth}{!}{%
\begin{tabular}{
@{}
l
!{\datasetsep}
>{\columncolor{UrbanBG}}c
>{\columncolor{UrbanBG}}c
>{\columncolor{UrbanBG}}c
>{\columncolor{UrbanBG}}c
>{\columncolor{BestBG}}c
!{\datasetsep}
>{\columncolor{JasperBG}}c
>{\columncolor{JasperBG}}c
>{\columncolor{JasperBG}}c
>{\columncolor{JasperBG}}c
>{\columncolor{BestBG}}c
!{\datasetsep}
>{\columncolor{StonewallBG}}c
>{\columncolor{StonewallBG}}c
>{\columncolor{StonewallBG}}c
>{\columncolor{StonewallBG}}c
>{\columncolor{BestBG}}c
@{}
}
\toprule

&
\multicolumn{5}{>{\columncolor{UrbanBG}}c}{\textbf{HYDICE Urban}} &
\multicolumn{5}{>{\columncolor{JasperBG}}c}{\textbf{Jasper Ridge}} &
\multicolumn{5}{>{\columncolor{StonewallBG}}c}{\textbf{Stonewall Playa}} \\

\cmidrule(lr){2-6}
\cmidrule(lr){7-11}
\cmidrule(l){12-16}

\textbf{Pipeline / method}
& $\Delta K\,\downarrow$
& mSAD $\downarrow$
& aRMSE $\downarrow$
& rRMSE $\downarrow$
& \cellcolor{BestBG}\textbf{Best}
& $\Delta K\,\downarrow$
& mSAD $\downarrow$
& aRMSE $\downarrow$
& rRMSE $\downarrow$
& \cellcolor{BestBG}\textbf{Best}
& $\Delta K\,\downarrow$
& mSAD $\downarrow$
& aRMSE $\downarrow$
& rRMSE $\downarrow$
& \cellcolor{BestBG}\textbf{Best} \\

\midrule


\rowcolor{UrbanLine!10}
\multicolumn{16}{@{}l}{%
    \textbf{HySime initialization}
} \\

\addlinespace[2pt]

CNN-AE \cite{Palsson2021CNNAEU}
& \textbf{1.9}
& 0.622
& \textbf{0.334}
& 0.182
& \textbf{2}
& 4.6
& 0.754
& \textbf{0.429}
& 0.089
& 1
& 7.0
& 0.221
& 0.370
& 0.105
& 0 \\

%

\addlinespace[1.5pt]

R-CoNMF \cite{DBLP:journals/tgrs/LiBPL16}
& 13.3
& 0.492
& 0.385
& 0.142
& 0
& 9.0
& 0.539
& 0.501
& 0.066
& 0
& 1.5
& \underline{0.198}
& \underline{0.300}
& \underline{0.035}
& 0 \\

\addlinespace[1.5pt]

uDAS \cite{Qu2019uDAS}
& 6.3
& 0.685
& \underline{0.350}
& 0.118
& 0
& 10.1
& 0.672
& 0.477
& \underline{0.059}
& 0
& 7.4
& 0.214
& 0.385
& 0.079
& 0 \\

%

\specialrule{0.45pt}{2.5pt}{2pt}

VCA--FCLSU
& 6.6
& \underline{0.448}
& 0.363
& \underline{0.097}
& 0
& 3.1
& \underline{0.524}
& 0.503
& \textbf{0.049}
& 1
& 2.3
& 0.199
& 0.378
& 0.064
& 0 \\

\addlinespace[1.5pt]

VCA--UnDIP
& 2.1
& \textbf{0.441}
& 0.376
& \textbf{0.077}
& \textbf{2}
& \underline{2.5}
& \textbf{0.502}
& 0.505
& \textbf{0.049}
& \textbf{2}
& \textbf{1.1}
& \textbf{0.197}
& \textbf{0.291}
& \textbf{0.033}
& \textbf{4} \\

\addlinespace[1.5pt]

SISAL--FCLSU
& \underline{2.0} & 0.507 & 0.352 & \underline{0.097}
& \cellcolor{BestBG}0
& 3.3
& 0.663
& \underline{0.441}
& 0.105
& 0
& \underline{1.2}
& 0.278
& 0.301
& 0.067
& 0 \\

\addlinespace[1.5pt]

SISAL--UnDIP
& 3.4
& 0.731
& 0.371
& 0.289
& 0
& \textbf{2.2}
& 0.699
& 0.481
& 0.098
& 1
& \textbf{1.1}
& 0.428
& 0.348
& 0.069
& 1 \\


\specialrule{0.80pt}{3pt}{0pt}

\rowcolor{black!7}
\multicolumn{16}{@{}l}{%
    \textbf{NWHFC initialization}
} \\

\addlinespace[2pt]

CNN-AE \cite{Palsson2021CNNAEU}
& \textbf{2.1}
& 0.577
& \underline{0.344}
& 0.172
& \textbf{1}
& 5.6
& 0.734
& 0.435
& 0.084
& 0
& 3.9
& 0.222
& 0.332
& 0.099
& 0 \\

%

\addlinespace[1.5pt]

R-CoNMF \cite{DBLP:journals/tgrs/LiBPL16}
& 17.1
& 0.484
& 0.387
& 0.149
& 0
& 7.1
& 0.541
& 0.491
& 0.053
& 0
& 2.6
& 0.204
& 0.292
& \underline{0.054}
& 0 \\

\addlinespace[1.5pt]

uDAS \cite{Qu2019uDAS}
& 8.9
& 0.665
& 0.359
& 0.140
& 0
& 8.0
& 0.669
& 0.471
& 0.054
& 0
& 7.4
& 0.214
& 0.385
& 0.079
& 0 \\

%

\specialrule{0.45pt}{2.5pt}{2pt}

VCA--FCLSU
& \underline{6.2}
& \textbf{0.451}
& 0.376
& 0.090
& \textbf{1}
& \textbf{1.3}
& \textbf{0.508}
& 0.484
& \underline{0.050}
& \textbf{2}
& 2.0
& \textbf{0.199}
& 0.332
& 0.068
& \textbf{1} \\

\addlinespace[1.5pt]

VCA--UnDIP
& 9.7
& \underline{0.482}
& 0.382
& \underline{0.089}
& 0
& \textbf{1.3}
& \underline{0.510}
& 0.499
& \textbf{0.042}
& \textbf{2}
& 2.2
& \underline{0.202}
& 0.396
& \textbf{0.045}
& \textbf{1} \\

\addlinespace[1.5pt]

SISAL--FCLSU
& 8.1
& 0.590
& \textbf{0.330}
& 0.105
& \textbf{1}
& 3.3
& 0.854
& \textbf{0.399}
& 0.122
& 1
& \underline{0.9}
& 0.551
& \textbf{0.231}
& 0.185
& \textbf{1} \\

\addlinespace[1.5pt]

SISAL--UnDIP
& 11.0
& 0.608
& 0.352
& \textbf{0.086}
& \textbf{1}
& \underline{2.5}
& 0.806
& \underline{0.421}
& 0.099
& 0
& \textbf{0.8}
& 0.569
& \underline{0.273}
& 0.149
& \textbf{1} \\

\bottomrule
\end{tabular}%
}
\end{table*}

\subsection{Results and Discussion}


\paragraph{To what extent does agent refinement improve the initial pipelines?}
The initial modular pipelines systematically overestimate the endmember cardinality, while the agent can only reduce the active set through merge and discard operations. The decrease in $\Delta K$ observed across all pipeline configurations is therefore partly favored by the one-sided action space and should not, by itself, be interpreted as evidence of a better decomposition. More importantly, the cardinality reductions are accompanied by lower mSAD in 22 of 24 comparisons, suggesting that the agent generally consolidates or removes redundant components while yielding more accurate spectral representatives. Improvements in aRMSE are less uniform, occurring in 16 of 24 comparisons, as the recovered abundance maps depend both on the refined endmember set and on the initial abundance estimation.
Despite using substantially fewer components, the refinement better explains the observed data, with rRMSE decreasing in 23 of 24 comparisons. Thus, the improvements in cardinality and spectral agreement are typically not obtained at the expense of reconstruction fidelity. Nevertheless, rRMSE remains complementary to other metrics, since an overcomplete decomposition may achieve low reconstruction error by using additional components to fit the observations.

\paragraph{How does the proposed approach compare with end-to-end unmixing methods?}
Table~\ref{tab:sota} compares the refined modular pipelines with representative integrated unmixing methods (CNN-AE~\cite{Palsson2021CNNAEU}, R-CoNMF~\cite{DBLP:journals/tgrs/LiBPL16}, and uDAS~\cite{Qu2019uDAS}). These baselines are functionally comparable to the proposed approach, as they start from an initial upper bound $K_{\max}$ and produce a complete unmixing solution comprising an active endmember set, its cardinality, and the corresponding abundance maps. To enable a direct comparison, the estimate by HySime or NWHFC is used as $K_0$ for the modular pipelines and as $K_{\max}$ for the integrated methods. 

At the initialization-group level, taking the best refined configuration for each metric, the NWHFC-initialized pipelines achieve the best mSAD, aRMSE, and rRMSE on all three datasets, as well as the best $\Delta K$ on Jasper Ridge and Stonewall Playa. Collectively, they attain the best result in 11 of the 12 dataset--metric comparisons. Under HySime initialization, the refined pipelines attain the best result in 9 of the 12 comparisons. Within this group, CNN-AE achieves the lowest aRMSE on HYDICE Urban and Jasper Ridge; however, the refined pipelines obtain substantially lower mSAD and rRMSE in both cases, indicating better spectral fidelity and reconstruction despite CNN-AE's advantage in abundance estimation. Although the metric-wise wins under NWHFC are distributed across multiple pipelines, HySime--VCA--UnDIP alone attains or shares the best result in 8 of the 12 comparisons, providing the strongest overall performance among the individual configurations. We therefore adopt HySime--VCA--UnDIP as the reference configuration for the subsequent ablations. These results complement the paired-pipeline analysis: the proposed agent not only improves on the initial decompositions but also produces unmixing solutions that are highly competitive with dedicated integrated approaches.

\paragraph{Computational demands and practical trade-offs.}
The proposed LVLM-driven framework is substantially more computationally demanding than conventional numerical unmixing pipelines because it requires repeated model inference, tool execution, and abundance re-estimation. For reference, on the more challenging HYDICE Urban dataset, Qwen 3.6 27B operating on a single H100 GPU needs around 25 minutes to complete the agentic inference, compared with, for instance, uDAS, which requires 15 minutes. The practical significance of this overhead, however, depends on the application. Our system provides a traceable unmixing process in which experts can inspect the acquired evidence, tool calls, and refinement actions, supporting human oversight in critical, non-latency-sensitive scenarios. In this sense, hyperspectral analysis can support resource-intensive activities such as geological prospecting and field surveys~\cite{Bedini2017Use}, cultural heritage characterization and conservation~\cite{Picollo2020HyperSpectral}, and forest pest detection and management~\cite{Kautz2024Early}. In such decision-support settings, computational cost may remain small relative to field operations, specialist labor, and interventions. The observed improvements in decomposition accuracy may therefore justify the additional computation in high-value offline analyses where traceability and expert oversight are more important than low latency. 

\begin{figure}[t]
    \centering
    \captionsetup{skip=3pt}
    \includegraphics[width=0.95\columnwidth]{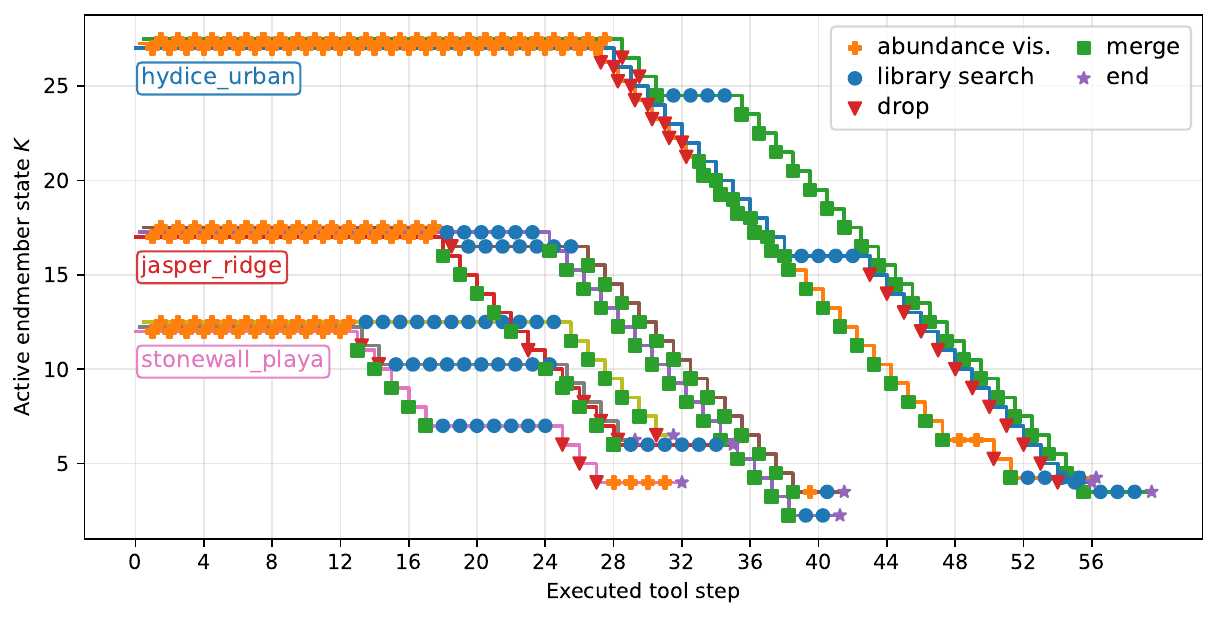}
    \caption{\textbf{Agent state trajectories.} Evolution of endmember count over executed tool steps for Qwen-3.6 27B with the HySime--VCA--UnDIP backbone. For each dataset, we report three independent runs. Markers identify specific operations.}
    \label{fig:agent-state-trajectories}
\end{figure}

\paragraph{Agent Trajectories Visualized.}
Figure~\ref{fig:agent-state-trajectories} illustrates representative state trajectories of the unmixing agent, reporting three different independent runs on each dataset. The horizontal segments show evidence-gathering steps that leave the active endmember count $K$ unchanged, whereas downward transitions correspond to merge or discard operations. Across the displayed runs, the agent first acts by inspecting the abundance maps before modifying the decomposition. In several trajectories, it then consults the material library before the first state-changing action; in others, library retrieval is deferred until after an initial refinement phase. Additional library queries or abundance-map visualizations may also be interleaved with later merge and discard operations, allowing the agent to reassess spectral and spatial evidence as the active endmember set evolves. 

\begin{table}[t]
\captionsetup{skip=4pt}
\centering
\footnotesize
\setlength{\tabcolsep}{3pt}
\setlength{\belowcaptionskip}{3pt}
\renewcommand{\arraystretch}{1.08}

\caption{\textbf{Controller ablation.}
HYDICE Urban with HySime--VCA--UnDIP initialization.}

\label{tab:ablation-decision-strategy}

\begin{tabular}{
@{}
l
!{\ablationsep}
cccc
@{}
}
\toprule

\textbf{Controller}
& $\Delta K\,\downarrow$
& mSAD $\downarrow$
& aRMSE $\downarrow$
& rRMSE $\downarrow$ \\

\midrule

Deterministic
& \underline{2.7}
& 0.553
& \textbf{0.368}
& \underline{0.096} \\

\addlinespace[1.5pt]

Ministral-3
& 8.2
& \underline{0.514}
& 0.389
& \underline{0.096} \\

\specialrule{0.6pt}{2.5pt}{1.5pt}

\textbf{Qwen-3.6 (adopted)}
& \textbf{2.1}
& \textbf{0.441}
& \underline{0.376}
& \textbf{0.077} \\

\bottomrule
\end{tabular}
\end{table}

\begin{table}[t]
\captionsetup{skip=4pt}
\centering
\footnotesize
\setlength{\tabcolsep}{3pt}
\setlength{\belowcaptionskip}{3pt}
\renewcommand{\arraystretch}{1.08}

\caption{\textbf{Agent-configuration ablation.}
HYDICE Urban with HySime--VCA--UnDIP initialization.}

\label{tab:ablation-agent-configurations}

\begin{tabular}{
@{}
l
!{\ablationsep}
cccc
@{}
}
\toprule

\textbf{Agent configuration}
& $\Delta K\,\downarrow$
& mSAD $\downarrow$
& aRMSE $\downarrow$
& rRMSE $\downarrow$ \\

\midrule

Library only (no images)
& 5.0
& 0.443
& \textbf{0.365}
& \underline{0.085} \\

\addlinespace[1.5pt]

Images only (no library)
& \textbf{1.7}
& \textbf{0.420}
& 0.392
& 0.102 \\

\specialrule{0.6pt}{2.5pt}{1.5pt}

\textbf{Library + images (adopted)}
& \underline{2.1}
& \underline{0.441}
& \underline{0.376}
& \textbf{0.077} \\

\bottomrule
\end{tabular}
\end{table}

\subsection{Ablation Study}
\label{sec:ablation}

\paragraph{Effect of the underlying controller.}
Table~\ref{tab:ablation-decision-strategy} compares the adopted
Qwen-3.6 strategy with Ministral-3 and a deterministic alternative, which drops and merges endmembers based on simple heuristics employing abundance support, dominant-pixel count, global abundance mass, and the reconstruction penalty induced by removing an endmember. Qwen-3.6 achieves the best $\Delta K$, mSAD, and rRMSE, while ranking second in aRMSE.
The deterministic strategy yields the lowest aRMSE but performs worse
on the remaining metrics, whereas Ministral-3 is consistently less
accurate than Qwen-3.6. 
performance.

\paragraph{Contribution of agent tools.}
Table~\ref{tab:ablation-agent-configurations} evaluates the spectral
library and abundance-map visualizations as complementary sources of
evidence. Using images alone yields the best cardinality and spectral
accuracy, whereas using the library alone achieves the lowest abundance
error. Combining both sources produces the best reconstruction accuracy
and ranks second on all remaining metrics. Thus, although neither source
dominates across all criteria, their combination provides the most
balanced performance. 

\section{Conclusion}

We introduced an algorithm-agnostic, LVLM-driven agentic framework for refining initial hyperspectral unmixing decompositions. Rather than replacing conventional unmixing algorithms, the proposed method operates on decompositions produced by modular pipelines and iteratively revises the active endmember set using complementary spectral and spatial evidence gathered through specialized tools. The results indicate that LVLM-guided refinement can improve the estimated endmember cardinality and the quality of the recovered spectral and abundance representations across different initializations, while remaining competitive with integrated end-to-end unmixing methods.

A current limitation is that refinement is restricted to merge and discard operations. Thus, the agent cannot add a new independent material component that is not represented in the initial decomposition. Extending the agent with mechanisms to propose new candidate endmembers, along with broader evaluation across scenes and underlying algorithms, is a promising direction for future work.

{
    \small
    \bibliographystyle{ieeenat_fullname}
    \bibliography{main}
}

\clearpage
\appendix
\setcounter{page}{1}


\twocolumn[
\begin{center}
    {\LARGE\bfseries Supplementary Material}
\end{center}
\vspace{0.5cm}
]

\section{Evaluation Metrics}
\label{sec:metrics}
We evaluate complete unmixing outputs against a reference decomposition
$(K_{\mathrm{gt}},\mathbf{M}_{\mathrm{gt}},\mathbf{A}_{\mathrm{gt}})$,
where $\mathbf{M}_{\mathrm{gt}}\in\mathbb{R}^{L\times K_{\mathrm{gt}}}$
and $\mathbf{A}_{\mathrm{gt}}\in\mathbb{R}^{K_{\mathrm{gt}}\times N}$.
For an estimated decomposition
$(\widehat K,\widehat{\mathbf{M}},\widehat{\mathbf{A}})$, the corresponding
matrices have dimensions $L\times\widehat K$ and $\widehat K\times N$,
respectively. We emphasize that these reference quantities are used only for post-hoc evaluation: no evaluated method receives the reference cardinality, spectral signatures, or abundance maps at inference time.

\paragraph{Cardinality error.}
For a run $r$, the endmember-count error is
\begin{equation}
    \Delta K^{(r)} =
    \left|\widehat K^{(r)}-K_{\mathrm{gt}}\right|.
\end{equation}
We report its mean absolute error across $R$ repeated runs,
\begin{equation}
    \Delta K = \frac{1}{R}\sum_{r=1}^{R}
    \left|\widehat K^{(r)}-K_{\mathrm{gt}}\right|.
\end{equation}
This measure penalizes both missed and spurious components, while allowing each
method to infer its cardinality directly from the observed scene.

\paragraph{Set-level spectral agreement.}
Let
\begin{equation}
    C_{ij} = \operatorname{SAD}
    \left(\mathbf{m}^{\mathrm{gt}}_i,\widehat{\mathbf{m}}_j\right)
\end{equation}
be the spectral angle distance between reference endmember $i$ and estimate
$j$~\cite{Keshava2002}. We compute the mean SAD (mSAD) over the complete
rectangular cost matrix,
\begin{equation}
    \operatorname{mSAD} =
    \frac{1}{K_{\mathrm{gt}}\widehat K}
    \sum_{i=1}^{K_{\mathrm{gt}}}\sum_{j=1}^{\widehat K} C_{ij}.
\end{equation}
Unlike correspondence-based or best-subset scores, this assignment-free quantity averages all reference--estimate relations, preventing poorly compatible estimated components from being hidden by a favorable subset of matches, as done in~\cite{s25082592}. It therefore provides a global measure of spectral compatibility when $\widehat K$ is not fixed \textit{a priori}.

\paragraph{Cardinality-aware abundance error.}
To compare estimated abundance maps with ground truth ones, we first compute a rectangular one-to-one Hungarian
correspondence using $C$ as the cost matrix~\cite{Kuhn1955}. With
$q=\min(K_{\mathrm{gt}},\widehat K)$, let
\begin{equation}
    \mathcal{H} = \arg\min_{\mathcal{P}\in\Pi_q}
    \sum_{(i,j)\in\mathcal{P}} C_{ij},
\end{equation}
where $\Pi_q$ is the set of size-$q$ one-to-one assignments. The unmatched
reference and estimated indices are denoted by
$\mathcal{U}_{\mathrm{gt}}$ and $\mathcal{U}_{\mathrm{est}}$, respectively.
Writing $\mathbf{a}^{\mathrm{gt}}_i$ and
$\widehat{\mathbf{a}}_j\in\mathbb{R}^{N}$ for abundance-map rows, we define
\begin{equation}
\begin{aligned}
    \operatorname{aRMSE} =
    \Bigg[\frac{1}{K_{\mathrm{gt}}N}
    \Bigg(&\sum_{(i,j)\in\mathcal{H}}
    \left\|\mathbf{a}^{\mathrm{gt}}_i-\widehat{\mathbf{a}}_j\right\|_2^2
    + \sum_{i\in\mathcal{U}_{\mathrm{gt}}}
    \left\|\mathbf{a}^{\mathrm{gt}}_i\right\|_2^2 \\
    &+ \sum_{j\in\mathcal{U}_{\mathrm{est}}}
    \left\|\widehat{\mathbf{a}}_j\right\|_2^2\Bigg)\Bigg]^{1/2}.
\end{aligned}
\end{equation}
Thus, unmatched components are compared with zero abundance maps. This construction (i) handles cardinality mismatch when $\widehat K\neq K_{\mathrm{gt}}$ and (ii) penalizes missing reference components and additional estimated components through their unmatched abundance mass.

\paragraph{Correspondence-conditioned reconstruction error.}
Let $\mathcal{J}=\{j:(i,j)\in\mathcal{H}\}$ be the estimated components
selected by the Hungarian correspondence. We retain only their abundance rows
and renormalize them at each pixel,
\begin{equation}
    \widetilde a_{jn} =
    \begin{cases}
        \displaystyle \frac{\widehat a_{jn}}
        {\sum_{\ell\in\mathcal{J}}\widehat a_{\ell n}}, &
        \displaystyle \sum_{\ell\in\mathcal{J}}\widehat a_{\ell n}>0,\\[8pt]
        0, & \text{otherwise},
    \end{cases}
    \qquad j\in\mathcal{J}.
\end{equation}
The reconstruction RMSE is then
\begin{equation}
    \operatorname{rRMSE} =
    \sqrt{\frac{1}{LN}
    \left\|\mathbf{Y} -
    \widehat{\mathbf{M}}_{\mathcal{J}}
    \widetilde{\mathbf{A}}_{\mathcal{J}}\right\|_F^2}.
\end{equation}
By reconstructing with only correspondence-selected components, rRMSE prevents
unmatched extra endmembers from improving the score through a larger model;
the per-pixel renormalization evaluates the selected components as a complete
mixture~\cite{Paura2023Benchmark}.

All four metrics are lower-is-better. Each metric is computed independently for every run, and the reported value is its mean across the $R$ repeated runs. Together, they assess cardinality recovery, global spectral compatibility, abundance fidelity under cardinality mismatch, and reconstruction quality using the components selected by the reference correspondence.

\section{Tool-Call Statistics}
We analyze how the agent uses its tools during refinement. Table~\ref{tab:agent-execution-statistics} summarizes the runs of Qwen-3.6 27B using the adopted weighted-average merge strategy across all eight modular pipelines. The agent requires $18.12$ iterations on average and performs $5.98$ discard and $6.81$ merge operations per run. HYDICE Urban produces substantially longer trajectories and more state changes than Jasper Ridge and Stonewall Playa, consistent with its larger initial candidate sets. Spectral-library retrieval remains frequent across all datasets, averaging $7.60$ accesses per run.

\begin{figure}[t]
    \centering
    \setlength{\abovecaptionskip}{5pt}




    \datasetrow
        {Samson}
        {SamsonLine}
        {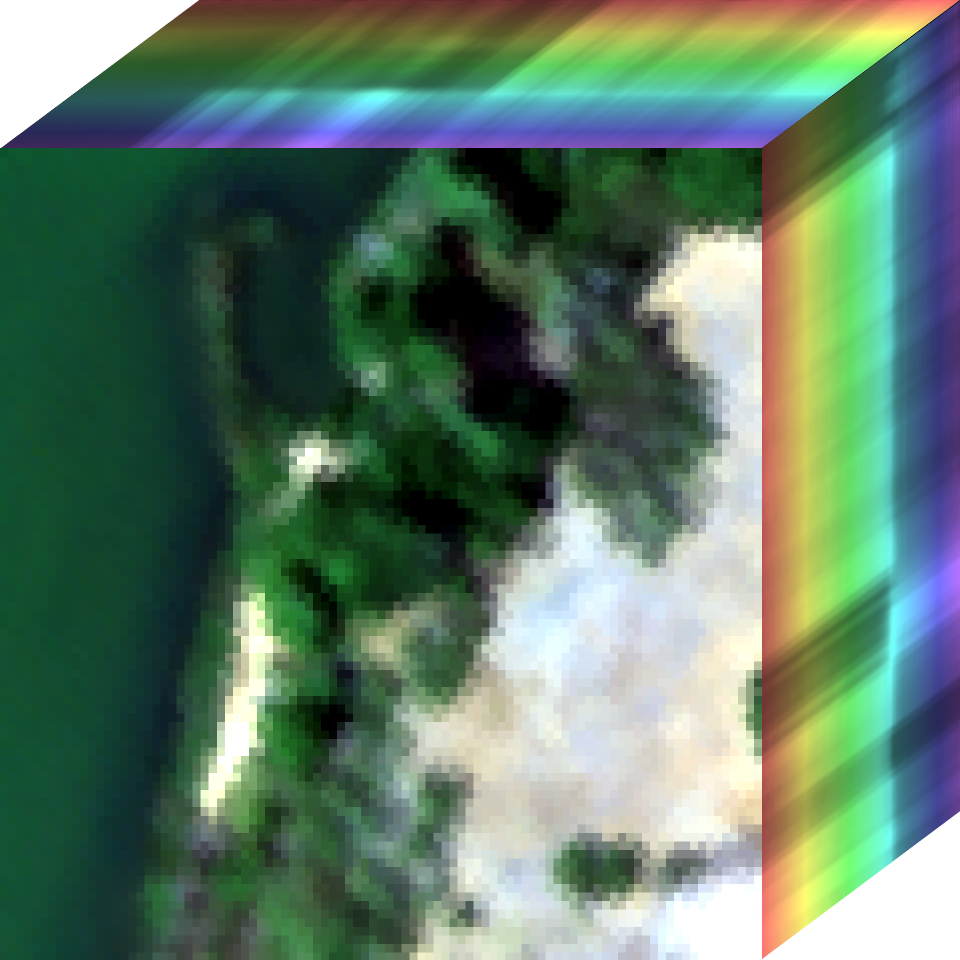}
        {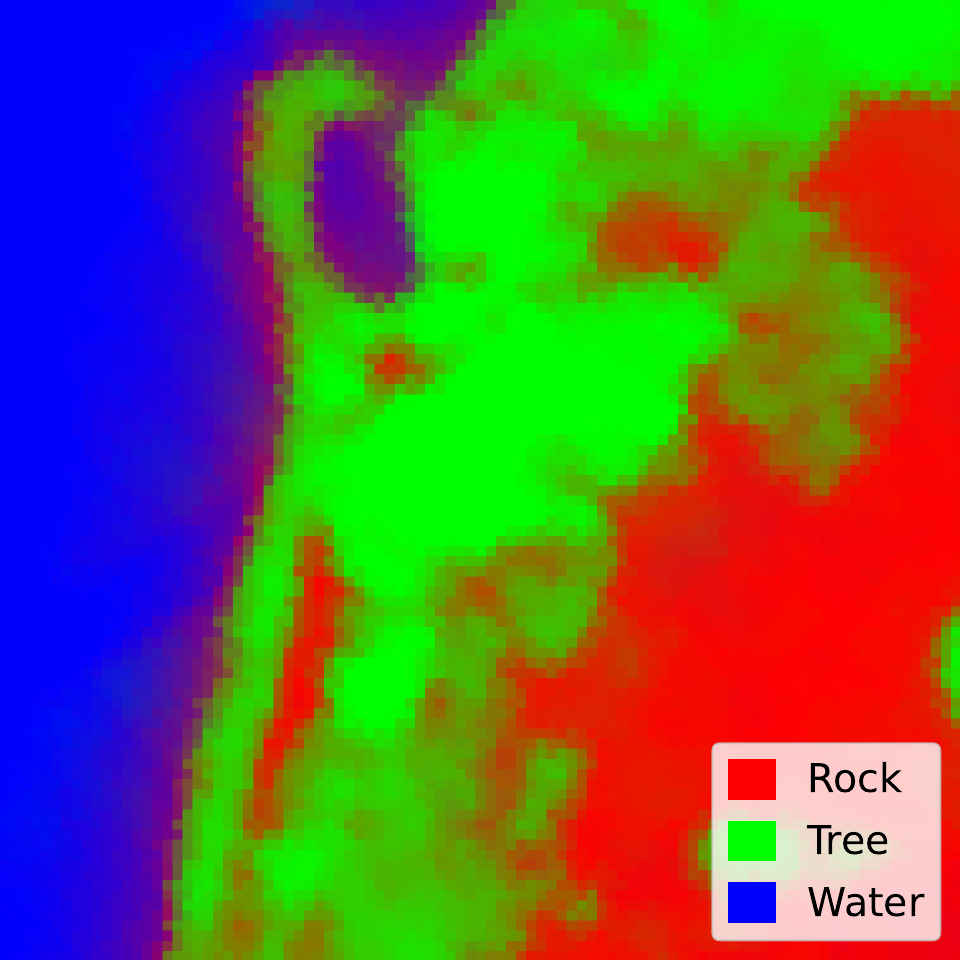}
        {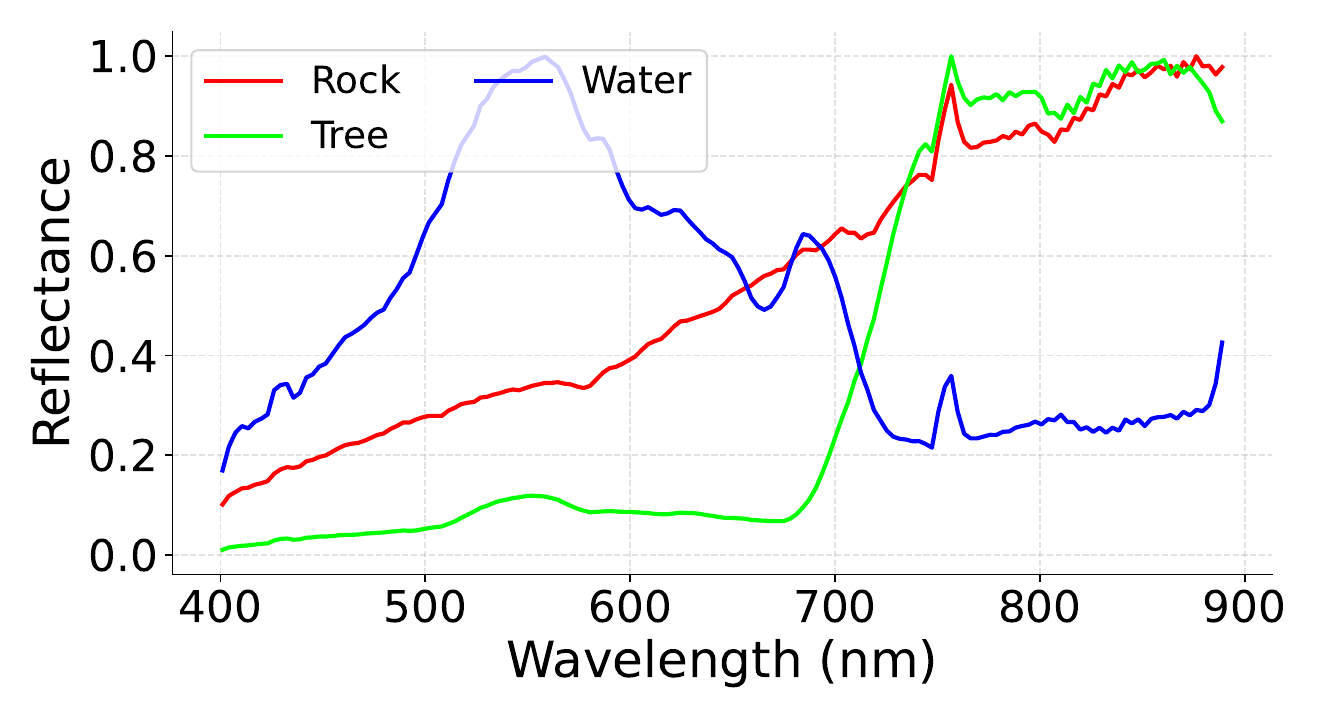}

    \vspace{.5mm}

    \caption{Overview of the Samson dataset. Each row
    shows the false-color image, abundance maps, and reference spectral
    signatures, from left to right.}

    \label{fig:datasets_overview_suppl}
\end{figure}

\section{Additional Results on Samson Dataset}

We report additional results for the Samson~\cite{zhu2017survey} dataset. Acquired using the Samson sensor, the original hyperspectral image contains \(952 \times 952\) pixels and 156 spectral bands spanning wavelengths from 0.401 to 0.889 $\mu$m. Most studies use a \(95 \times 95\)-pixel subscene, resulting in a hyperspectral cube of \(95 \times 95 \times 156\) (Figure~\ref{fig:datasets_overview_suppl}). The scene comprises three endmembers—--soil, trees, and water.

\paragraph{Agentic refinement of the initial modular pipelines.}
Table~\ref{tab:initial-agent-samson} demonstrates that the proposed agent substantially improves the initial modular decompositions on Samson, particularly in endmember-count and abundance estimation. Under HySime, refinement reduces $\Delta K$ from $40$ to $8.0$--$17.4$ and improves aRMSE for all four backbones. Under the more favorable NWHFC initialization, it further reduces $\Delta K$ from $5$ to approximately one endmember and consistently lowers both aRMSE and rRMSE. The strongest results are obtained with VCA--UnDIP, which reaches $\Delta K=0.7$, aRMSE $=0.304$, and rRMSE $=0.041$. The agent also improves mSAD for all SISAL-based decompositions, confirming its ability to refine heterogeneous initial solutions rather than relying on a single numerical backbone.

\paragraph{Comparison with end-to-end unmixing methods.}
Table~\ref{tab:sota-samson} shows that the refined pipelines remain competitive with dedicated end-to-end unmixing methods. Under HySime, the proposed approach achieves the best spectral and reconstruction results: VCA--FCLSU obtains the lowest mSAD ($0.508$), while VCA--UnDIP attains the lowest rRMSE ($0.075$). Its advantage becomes clearer under NWHFC, where the refined pipelines collectively win all four metrics. In particular, VCA--UnDIP achieves the best $\Delta K$ ($0.7$), aRMSE ($0.304$), and rRMSE ($0.041$), while VCA--FCLSU provides the best mSAD ($0.536$). Nevertheless, Samson also exposes two limitations. When HySime severely overestimates the model order, the agent substantially reduces but does not fully eliminate the cardinality error, and CNN-AE remains better in $\Delta K$ and aRMSE. Moreover, the weaker SISAL results show that refinement remains dependent on the quality of the initial candidate signatures and cannot fully recover from an unfavorable underlying decomposition.

\begin{table}[t]
\centering
\footnotesize
\setlength{\tabcolsep}{3pt}
\setlength{\belowcaptionskip}{3pt}
\renewcommand{\arraystretch}{1.08}

\caption{\textbf{Agent execution statistics by dataset.}
Iterations, discards, merges, and library accesses are averaged per run.
The Overall row gives each dataset equal weight.}

\label{tab:agent-execution-statistics}

\begin{tabular}{
@{}
l
!{\ablationsep}
cccc
@{}
}
\toprule

\textbf{Dataset}
& \textbf{Iter.}
& \textbf{Drop}
& \textbf{Merge}
& \textbf{Library} \\

\midrule

Jasper Ridge
& 18.19
& 3.31
& 6.55
& 8.15 \\

\addlinespace[1.5pt]

HYDICE Urban
& 25.43
& 12.33
& 10.54
& 6.91 \\

\addlinespace[1.5pt]

Stonewall Playa
& 10.75
& 2.31
& 3.34
& 7.74 \\

\specialrule{0.6pt}{2.5pt}{1.5pt}

\textbf{Overall}
& \textbf{18.12}
& \textbf{5.98}
& \textbf{6.81}
& \textbf{7.60} \\

\bottomrule
\end{tabular}
\end{table}

\begin{table}[t]
\centering
\footnotesize
\setlength{\tabcolsep}{3pt}
\renewcommand{\arraystretch}{1.05}

\caption{\textbf{Initial and agent-refined modular decompositions on
Samson.}
For each pipeline, we report the initial decomposition and its
agent-refined counterpart. The better value within each pair is bolded;
lower is better for all metrics.}

\label{tab:initial-agent-samson}

\begin{adjustbox}{max width=\columnwidth}
\begin{tabular}{
@{}
l
!{\datasetsep}
l
!{\datasetsep}
>{\columncolor{SamsonBG}}c
>{\columncolor{SamsonBG}}c
>{\columncolor{SamsonBG}}c
>{\columncolor{SamsonBG}}c
@{}
}
\toprule

\textbf{Pipeline}
& \textbf{Stage}
& $\Delta K\,\downarrow$
& mSAD $\downarrow$
& aRMSE $\downarrow$
& rRMSE $\downarrow$ \\

\midrule

\multirow{2}{*}{HySime--VCA--FCLSU}
& Initial
& 40.0
& \textbf{0.498}
& 0.505
& 0.190 \\

& + Agent
& \textbf{8.0}
& 0.508
& \textbf{0.433}
& \textbf{0.130} \\

\addlinespace[1.8pt]

\multirow{2}{*}{HySime--VCA--UnDIP}
& Initial
& 40.0
& \textbf{0.498}
& 0.506
& 0.079 \\

& + Agent
& \textbf{13.5}
& 0.527
& \textbf{0.414}
& \textbf{0.075} \\

\addlinespace[1.8pt]

\multirow{2}{*}{HySime--SISAL--FCLSU}
& Initial
& 40.0
& 0.779
& 0.497
& \textbf{0.146} \\

& + Agent
& \textbf{17.4}
& \textbf{0.685}
& \textbf{0.455}
& 0.196 \\

\addlinespace[1.8pt]

\multirow{2}{*}{HySime--SISAL--UnDIP}
& Initial
& 40.0
& 0.779
& 0.537
& \textbf{0.183} \\

& + Agent
& \textbf{12.4}
& \textbf{0.703}
& \textbf{0.508}
& 0.206 \\

\addlinespace[1.8pt]

\multirow{2}{*}{NWHFC--VCA--FCLSU}
& Initial
& 5.0
& \textbf{0.534}
& 0.455
& 0.135 \\

& + Agent
& \textbf{1.0}
& 0.536
& \textbf{0.320}
& \textbf{0.066} \\

\addlinespace[1.8pt]

\multirow{2}{*}{NWHFC--VCA--UnDIP}
& Initial
& 5.0
& \textbf{0.534}
& 0.475
& 0.089 \\

& + Agent
& \textbf{0.7}
& 0.557
& \textbf{0.304}
& \textbf{0.041} \\

\addlinespace[1.8pt]

\multirow{2}{*}{NWHFC--SISAL--FCLSU}
& Initial
& 5.0
& 0.805
& 0.458
& 0.328 \\

& + Agent
& \textbf{1.1}
& \textbf{0.777}
& \textbf{0.424}
& \textbf{0.222} \\

\addlinespace[1.8pt]

\multirow{2}{*}{NWHFC--SISAL--UnDIP}
& Initial
& 5.0
& 0.805
& 0.499
& 0.326 \\

& + Agent
& \textbf{1.0}
& \textbf{0.771}
& \textbf{0.469}
& \textbf{0.175} \\

\bottomrule
\end{tabular}
\end{adjustbox}
\end{table}

\section{Additional Ablations}
\paragraph{Effect of the merge strategy.}
The merge tool may highly depend on the merge algorithm used to fuse the endmembers. In this section, we therefore explore the effect of the employed merge strategy by considering three options: \textit{average}, \textit{weighted average}, and \textit{medoid}. For each cluster of endmembers identified as mergeable, the average uses the arithmetic mean of all spectra in the cluster. Different from plain average, the weighted average weights each spectrum by its total estimated abundance across the image, giving greater influence to endmembers that explain more of the scene. Finally, the medoid approach selects the cluster member with the lowest spectrum with respect to the cluster mean.
Table~\ref{tab:ablation-merge-strategies} compares the three strategies. Simple averaging achieves the lowest cardinality error,
whereas the medoid yields the best abundance and reconstruction
accuracy. Weighted averaging obtains the lowest spectral error and
ranks second on both RMSE measures, although at the cost of a higher
$\Delta K$. We adopt weighted averaging because it prioritizes
endmember fidelity while retaining competitive abundance and
reconstruction accuracy.

\begin{table}[t]
\centering
\footnotesize
\setlength{\tabcolsep}{2.7pt}
\renewcommand{\arraystretch}{1.07}

\caption{\textbf{Comparison under HySime and NWHFC initialization
on Samson.}
Agent-refined modular pipelines and competing methods are evaluated
separately under each initialization. Within each group, best results
are bolded, second-best results are underlined, and Best counts the
metric-wise wins.}

\label{tab:sota-samson}

\begin{adjustbox}{max width=\columnwidth}
\begin{tabular}{
@{}
l
!{\datasetsep}
>{\columncolor{SamsonBG}}c
>{\columncolor{SamsonBG}}c
>{\columncolor{SamsonBG}}c
>{\columncolor{SamsonBG}}c
>{\columncolor{BestBG}}c
@{}
}
\toprule

\textbf{Pipeline / method}
& $\Delta K\,\downarrow$
& mSAD $\downarrow$
& aRMSE $\downarrow$
& rRMSE $\downarrow$
& \cellcolor{BestBG}\textbf{Best} \\

\midrule


\rowcolor{UrbanLine!10}
\multicolumn{6}{@{}l}{%
    \textbf{HySime initialization}
} \\

\addlinespace[1.5pt]

CNN-AE
& \textbf{0.0}
& 0.648
& \textbf{0.179}
& 0.147
& \textbf{2} \\

%

\addlinespace[1.2pt]

R-CoNMF
& \underline{4.9}
& 0.549
& \underline{0.385}
& \underline{0.100}
& 0 \\

\addlinespace[1.2pt]

UDAS
& 16.6
& 0.654
& 0.497
& 0.177
& 0 \\

%

\specialrule{0.45pt}{2.5pt}{2pt}

VCA--FCLSU
& 8.0
& \textbf{0.508}
& 0.433
& 0.130
& 1 \\

\addlinespace[1.2pt]

VCA--UnDIP
& 13.5
& \underline{0.527}
& 0.414
& \textbf{0.075}
& 1 \\

\addlinespace[1.2pt]

SISAL--FCLSU
& 17.4
& 0.685
& 0.455
& 0.196
& 0 \\

\addlinespace[1.2pt]

SISAL--UnDIP
& 12.4
& 0.703
& 0.508
& 0.206
& 0 \\


\specialrule{0.80pt}{3pt}{0pt}

\rowcolor{black!7}
\multicolumn{6}{@{}l}{%
    \textbf{NWHFC initialization}
} \\

\addlinespace[1.5pt]

CNN-AE
& 3.8
& 0.658
& 0.391
& 0.175
& 0 \\

%

\addlinespace[1.2pt]

R-CoNMF
& 2.9
& \underline{0.543}
& 0.421
& 0.089
& 0 \\

\addlinespace[1.2pt]

UDAS
& 5.0
& 0.589
& 0.433
& 0.105
& 0 \\

%

\specialrule{0.45pt}{2.5pt}{2pt}

VCA--FCLSU
& \underline{1.0}
& \textbf{0.536}
& \underline{0.320}
& \underline{0.066}
& 1 \\

\addlinespace[1.2pt]

VCA--UnDIP
& \textbf{0.7}
& 0.557
& \textbf{0.304}
& \textbf{0.041}
& \textbf{3} \\

\addlinespace[1.2pt]

SISAL--FCLSU
& 1.1
& 0.777
& 0.424
& 0.222
& 0 \\

\addlinespace[1.2pt]

SISAL--UnDIP
& \underline{1.0}
& 0.771
& 0.469
& 0.175
& 0 \\

\bottomrule
\end{tabular}
\end{adjustbox}
\end{table}

\begin{table}[t]
\centering
\footnotesize
\setlength{\tabcolsep}{3pt}
\setlength{\belowcaptionskip}{3pt}
\renewcommand{\arraystretch}{1.08}

\caption{\textbf{Merge-strategy ablation.}
HYDICE Urban with HySime--VCA--UnDIP initialization.}

\label{tab:ablation-merge-strategies}

\begin{tabular}{
@{}
l
!{\ablationsep}
cccc
@{}
}
\toprule

\textbf{Merge strategy}
& $\Delta K\,\downarrow$
& mSAD $\downarrow$
& aRMSE $\downarrow$
& rRMSE $\downarrow$ \\

\midrule

Average
& \textbf{1.4}
& \underline{0.464}
& 0.392
& 0.087 \\

\addlinespace[1.5pt]

Medoid
& \underline{1.8}
& 0.488
& \textbf{0.359}
& \textbf{0.070} \\

\specialrule{0.6pt}{2.5pt}{1.5pt}

\textbf{Weighted average (adopted)}
& 2.1
& \textbf{0.441}
& \underline{0.376}
& \underline{0.077} \\

\bottomrule
\end{tabular}
\end{table}

\section{Qualitative Analysis of Recovered Endmembers}
\begin{figure*}[t]
    \centering
    \begin{subfigure}{\textwidth}
        \centering
        \includegraphics[width=0.7\linewidth]{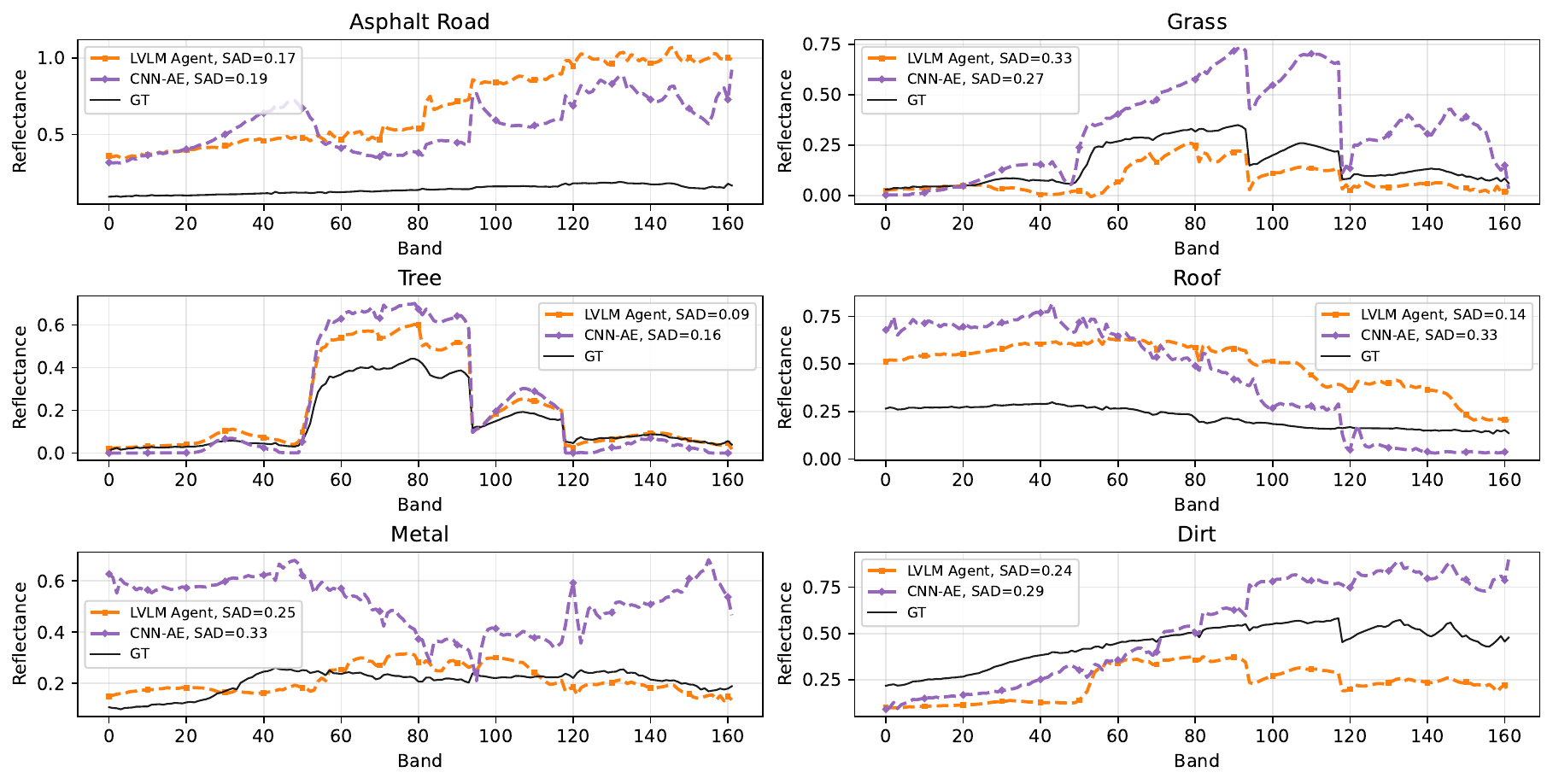}
        \caption{HYDICE Urban}
    \end{subfigure} 
    \begin{subfigure}{\textwidth}
        \centering
        \includegraphics[width=0.7\linewidth]{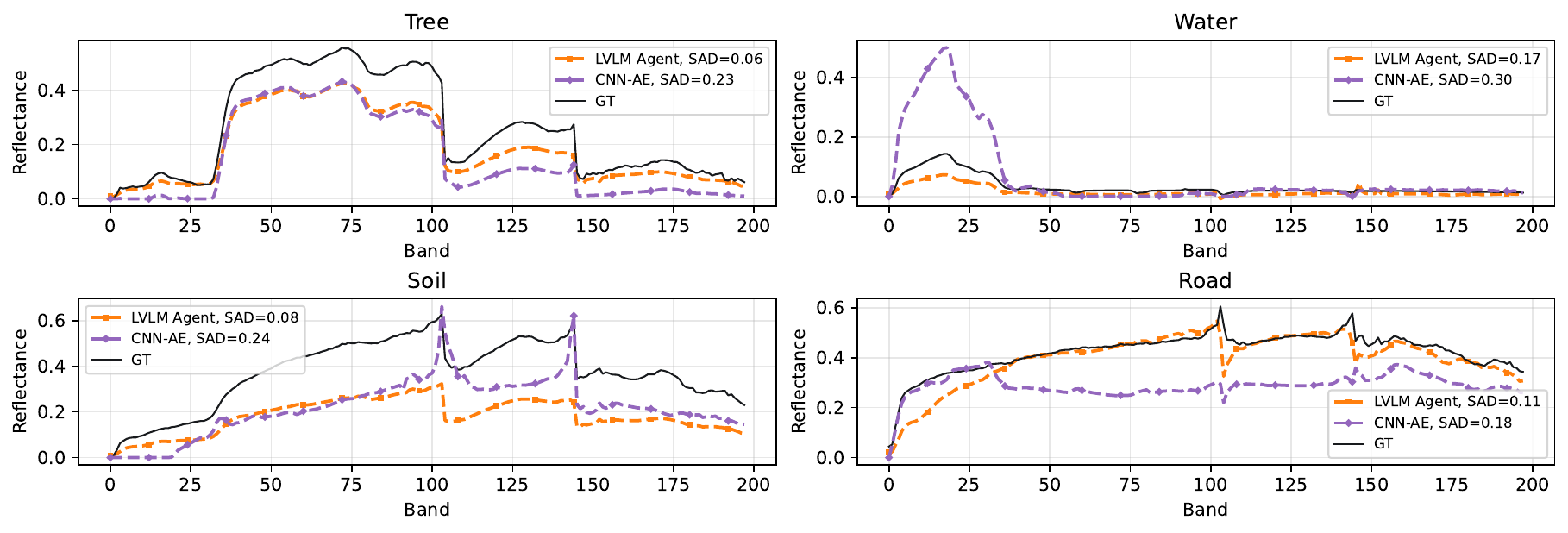}
        \caption{Jasper Ridge}
    \end{subfigure} 
    \begin{subfigure}{\textwidth}
        \centering
        \includegraphics[width=0.7\linewidth]{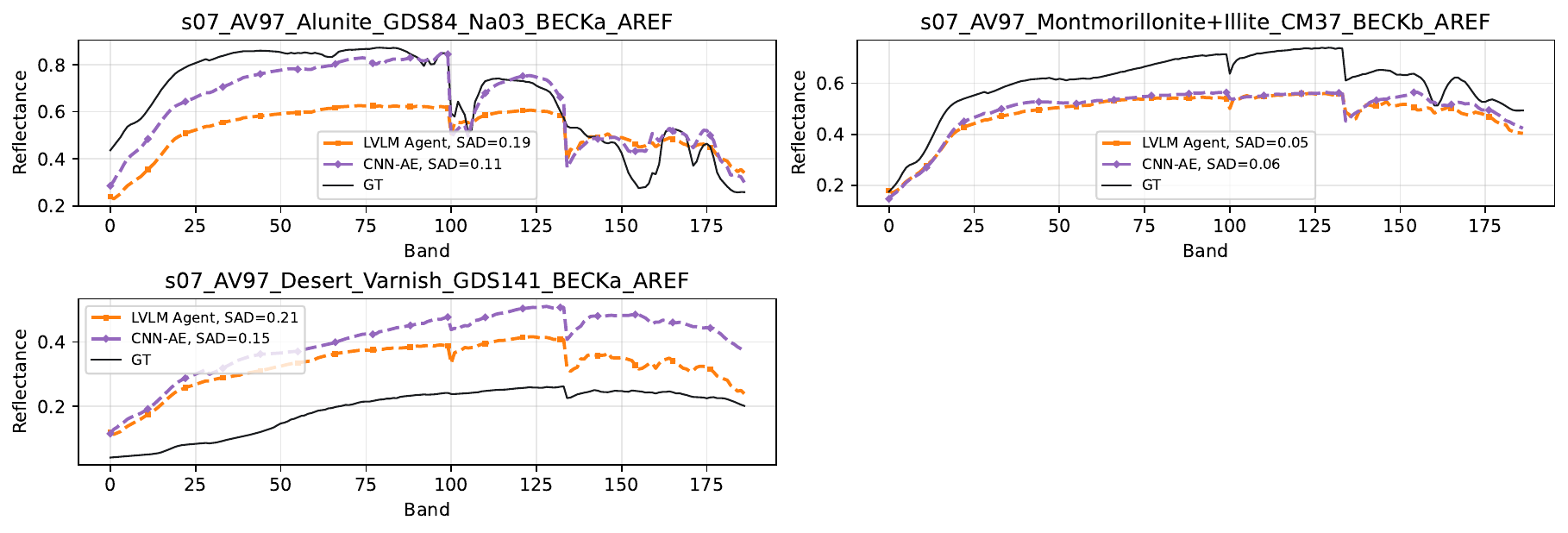}
        \caption{Stonewall Playa}
    \end{subfigure}
    \begin{subfigure}{\textwidth}
        \centering
        \includegraphics[width=0.7\linewidth]{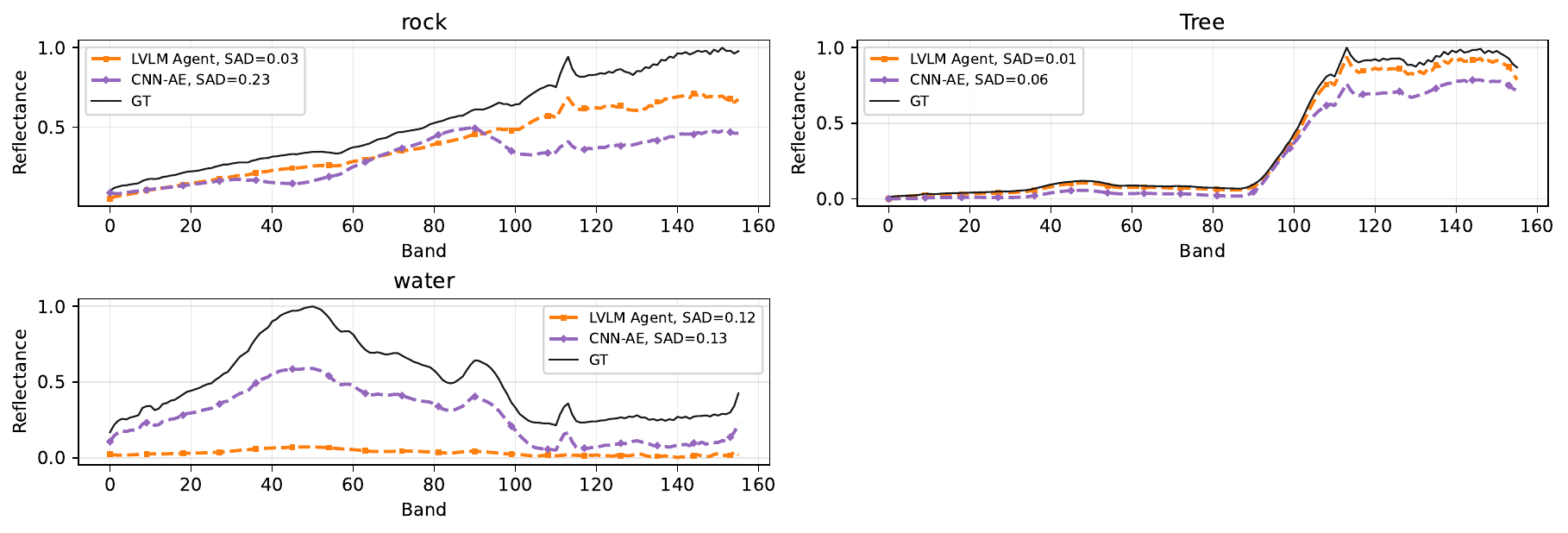}
        \caption{Samson}
    \end{subfigure}

\caption{\textbf{Qualitative comparison of recovered endmember spectra on HYDICE Urban, Jasper Ridge, Stonewall Playa, and Samson.} For each material, the reference spectrum is shown alongside the final signatures recovered by the adopted HySime--VCA--UnDIP LVLM agentic configuration and CNN-AE. The legends report the SAD to the corresponding reference signature; lower is better.}
\label{fig:em_spectra}
\end{figure*}

Figure~\ref{fig:em_spectra} compares the final endmember signatures recovered by our reference HySime--VCA--UnDIP agentic configuration and CNN-AE with the corresponding ground-truth spectra. CNN-AE is selected as a representative strong end-to-end method. The LVLM agent achieves a lower spectral angle distance for 13 of the 16 reference endmembers: five of six on HYDICE Urban, all four on Jasper Ridge, one of three on Stonewall Playa, and all three on Samson. On HYDICE Urban, the improvements are particularly evident for tree, roof, and metal, while CNN-AE performs better only for grass. The agent also provides consistently closer signatures on Jasper Ridge and Samson. Stonewall Playa is more mixed: the agent performs better for Montmorillonite/Illite, whereas CNN-AE obtains lower errors for Alunite and desert varnish. Overall, the comparison indicates that the refined signatures generally preserve the shapes of the reference spectra more accurately, although the relative performance remains material- and dataset-dependent.

\section{Detailed System Prompt and Agent Outputs}
\label{sec:qualitative_hydice}

In the following section, we illustrate the reasoning process of the proposed agentic pipeline when applied to the HYDICE Urban dataset. To provide a transparent view of the agent's decision-making behavior, we trace a single execution run from initialization to the final material classification, reporting output examples of the main tool calls. 

\begin{figure*}[t]
\centering
\begin{tcolorbox}[colback=gray!5!white, colframe=black!75, title={System Message: Agent Instructions}, fonttitle=\bfseries, boxsep=0pt, left=4pt, right=4pt, top=4pt, bottom=4pt]
\scriptsize

\textbf{[SystemMessage]} \\
You are an expert hyperspectral image analyst with deep knowledge of remote sensing materials, spectral reflectance physics, and geospatial interpretation.
Your task is to identify the physical materials present in a hyperspectral image (HSI) by reasoning over the evidence available in this configured ablation run.

\vspace{0.5ex}
\textbf{AVAILABLE TOOLS}
\hrule
\vspace{0.5ex}
\begin{enumerate}
    \setlength{\itemsep}{0pt} \setlength{\parskip}{0pt}
    \item \texttt{library\_search} - Find the closest real-world materials (by SAD).
    \item \texttt{compute\_abundance} - Visualise the abundance heatmap (image injected next turn).
    \item \texttt{merge\_endmembers} - Merge two endmembers that represent the same material.
    \item \texttt{discard\_endmember} - Remove a noisy or artifact endmember.
\end{enumerate}

\vspace{0.5ex}
\textbf{INTERNAL STATE TABLE} (update your mental model after every tool call)
\hrule
\vspace{0.5ex}
\begin{center}
\renewcommand{\arraystretch}{0.85}
\begin{tabular}{|l|l|l|l|l|}
\hline
\textbf{ID} & \textbf{Candidate Materials} & \textbf{Cluster Mass} & \textbf{Associated Image} & \textbf{Status} \\ \hline
E0 & ["Unknown"] & 0.342 & N/A & ACTIVE \\ \hline
... & ... & ... & ... & ... \\ \hline
\end{tabular}
\end{center}

\vspace{0.5ex}
\textbf{REASONING WORKFLOW}
\hrule
\vspace{0.5ex}
\begin{enumerate}
    \setlength{\itemsep}{0pt} \setlength{\parskip}{0pt}
    \item Initial decomposition has already been run before your first turn. Read the provided bootstrap JSON/table as the current endmember state.
    \item Use \texttt{compute\_abundance} when spatial evidence is needed for an endmember. Analyse the overlay image injected after each call:
    \begin{itemize}
        \setlength{\itemsep}{0pt} \setlength{\parskip}{0pt}
        \item \textbf{SPATIAL COHESION}: Does the heatmap highlight clean geometric shapes (roofs, roads) or contiguous organic zones (fields, canopies)? If the heatmap looks like salt-and-pepper noise uniformly spread across the image it is a strong candidate for discard or merge.
        \item \textbf{MASS GATEKEEPING}: If \texttt{cluster\_mass} < 0.005, investigate carefully. Discard if it represents noise; keep if it has spatial cohesion.
        \item \textbf{REDUNDANCY}: If two endmembers produce highly overlapping heatmaps, merge them with \texttt{merge\_endmembers}.
    \end{itemize}
    \item Call \texttt{library\_search} for each surviving endmember to gather spectral material hypotheses. Treat library matches as evidence, not ground truth.
    \item \textbf{SEMANTIC ALIGNMENT}: If \texttt{library\_search} returns a material that conflicts with the abundance overlay, reject or downgrade that identity hypothesis.
    \item Apply \texttt{merge\_endmembers} or \texttt{discard\_endmember} to refine the endmember set.
    \item Once no further merges or discards are warranted, produce a \textbf{FINAL REPORT}:
    \begin{itemize}
        \setlength{\itemsep}{0pt} \setlength{\parskip}{0pt}
        \item List every ACTIVE endmember with its identified material or `Unknown' and cluster mass.
        \item Summarise any notable scene observations supported by available evidence.
        \item Clearly state ``ANALYSIS COMPLETE'' at the end.
    \end{itemize}
\end{enumerate}

\vspace{0.5ex}
\textbf{CRITICAL RULES}
\hrule
\vspace{0.5ex}
\begin{itemize}
    \setlength{\itemsep}{0pt} \setlength{\parskip}{0pt}
    \item Base every decision on BOTH spectral evidence (library SAD) AND spatial evidence (abundance overlay).
    \item Do NOT discard endmembers solely because of low cluster mass if they show clear spatial cohesion.
    \item Stop only when no further merges or discards are warranted.
    \item When all reasonable actions have been exhausted, write the final report and end the response with the exact phrase ``ANALYSIS COMPLETE''.
\end{itemize}

The original RGB image of the scene is shown below.
\begin{center}
    \includegraphics[width=0.2\linewidth]{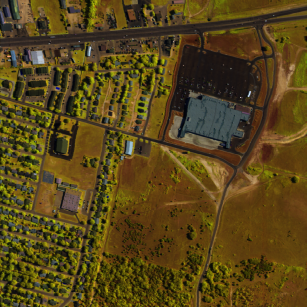} 
\end{center}

\end{tcolorbox}
\caption{}
\label{fig:prompt}
\end{figure*}

\begin{figure*}[t]
\centering
\begin{tcolorbox}[colback=gray!5!white, colframe=black!75, title={Endmember Analysis Interaction: E23 (Vegetation)}, fonttitle=\bfseries]
\scriptsize

\textbf{AI:} Now let's check \textbf{E23} (mass 0.2406):

\vspace{1ex}
\hrule
\vspace{1ex}

\textbf{[ToolMessage]} \\
\texttt{\{} \\
\texttt{~~"image\_path": "overlay\_E23.png",} \\
\texttt{~~"label": "E23",} \\
\texttt{~~"message": "Abundance overlay image attached to next turn context."} \\
\texttt{\}}

\vspace{1ex}
\hrule
\vspace{1ex}

\textbf{[HumanMessage]} \\
{[}Abundance overlay for `E23' is shown below. Analyse the spatial distribution of this endmember in the scene.{]}

\vspace{0.5em}
\begin{center}
    \includegraphics[width=0.4\linewidth]{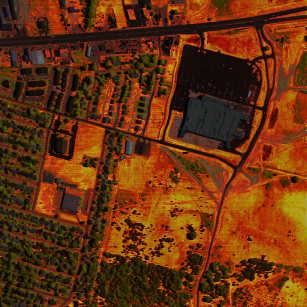} 
\end{center}
\vspace{0.5em}

\vspace{1ex}
\hrule
\vspace{1ex}

\textbf{[AIMessage]} \\
The abundance map for \textbf{E23} shows a strong signal on the vegetation (trees and grass) throughout the scene, particularly in the residential areas and open green spaces. The spectral vector for E23 shows low reflectance in the visible range and higher reflectance in the NIR/SWIR, which is characteristic of healthy vegetation. This appears to be a valid endmember representing vegetation.

\end{tcolorbox}
\caption{}
\label{fig:abund_1}
\end{figure*}

\begin{figure*}[t]
\centering
\begin{tcolorbox}[colback=gray!5!white, colframe=black!75, title={Endmember Analysis Interaction: E18 (Soil/Ground)}, fonttitle=\bfseries]
\scriptsize

\textbf{AI:} Now let's check \textbf{E18} (mass 0.1047):

\vspace{1ex}
\hrule
\vspace{1ex}

\textbf{[ToolMessage]} \\
\texttt{\{} \\
\texttt{~~"image\_path": "overlay\_E18.png",} \\
\texttt{~~"label": "E18",} \\
\texttt{~~"message": "Abundance overlay image attached to next turn context."} \\
\texttt{\}}

\vspace{1ex}
\hrule
\vspace{1ex}

\textbf{[HumanMessage]} \\
{[}Abundance overlay for `E18' is shown below. Analyse the spatial distribution of this endmember in the scene.{]}

\vspace{0.5em}
\begin{center}
    \includegraphics[width=0.4\linewidth]{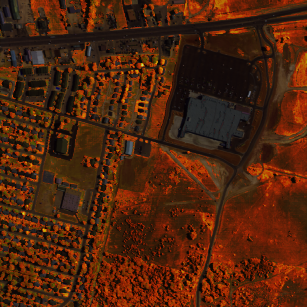} 
\end{center}
\vspace{0.5em}

\vspace{1ex}
\hrule
\vspace{1ex}

\textbf{[AIMessage]} \\
The abundance map for \textbf{E18} shows a strong signal on the bare soil/ground areas, particularly in the open fields and around the buildings. The spectral vector for E18 shows low reflectance in the visible range and moderate reflectance in the NIR/SWIR, which is characteristic of dry soil or dirt. This appears to be a valid endmember representing soil/dirt.

\end{tcolorbox}
\caption{}
\label{fig:abund_2}
\end{figure*}

\begin{figure*}[t]
\centering
\begin{tcolorbox}[colback=gray!5!white, colframe=black!75, title={Endmember Analysis Interaction: E0 (Noise/Artifact)}, fonttitle=\bfseries]
\scriptsize

\textbf{AI:} Now let's check some of the lower mass endmembers to see if they are noise or represent distinct materials. Let's start with \textbf{E0} (mass 0.0005), which has very low mass:

\vspace{1ex}
\hrule
\vspace{1ex}

\textbf{[ToolMessage]} \\
\texttt{\{} \\
\texttt{~~"image\_path": "overlay\_E0.png",} \\
\texttt{~~"label": "E0",} \\
\texttt{~~"message": "Abundance overlay image attached to next turn context."} \\
\texttt{\}}

\vspace{1ex}
\hrule
\vspace{1ex}

\textbf{[HumanMessage]} \\
{[}Abundance overlay for `E0' is shown below. Analyse the spatial distribution of this endmember in the scene.{]}

\vspace{0.5em}
\begin{center}
    \includegraphics[width=0.4\linewidth]{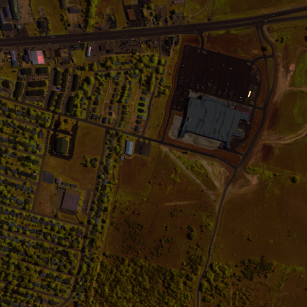} 
\end{center}
\vspace{0.5em}

\vspace{1ex}
\hrule
\vspace{1ex}

\textbf{[AIMessage]} \\
The abundance map for \textbf{E0} shows very little signal, with only a few scattered pixels highlighted. This appears to be noise or an artifact. Given its very low cluster mass (0.0005) and lack of spatial cohesion, I will discard this endmember.

\end{tcolorbox}
\caption{}
\label{fig:abund_3}
\end{figure*}

First, Figure~\ref{fig:prompt} details the initial system prompt, which establishes the agent's role, provides the reasoning workflow, and outlines the available tools. Once initialized, the agent iteratively evaluates the spectral and spatial properties of the candidate endmembers. Figures~\ref{fig:abund_1},\ref{fig:abund_2},\ref{fig:abund_3} showcase examples of the agent's abundance analysis, demonstrating how it leverages spatial cohesion and mass gatekeeping to validate reasonable materials (e.g., vegetation and soil) and discard noise artifacts. 

\begin{figure*}[t]
\centering
\begin{tcolorbox}[colback=gray!5!white, colframe=black!75, title={Final Report Generation}, fonttitle=\bfseries, boxsep=0pt, left=4pt, right=4pt, top=4pt, bottom=4pt]
\scriptsize

\textbf{[AIMessage]} \\
Now I have a clean set of active endmembers representing the main materials in the scene. Let me produce the final report.

\vspace{1ex}
\textbf{FINAL REPORT}

\vspace{0.5ex}
\textbf{Active Endmembers and Identified Materials}
\vspace{0.5ex}

\begin{center}
\renewcommand{\arraystretch}{1.1}
\begin{tabular}{|l|c|p{0.22\linewidth}|p{0.5\linewidth}|}
\hline
\textbf{ID} & \textbf{Cluster Mass} & \textbf{Identified Material} & \textbf{Description} \\
\hline
\textbf{E11} & 0.3747 & Dark Surfaces (Asphalt/Roofing) & Represents dark man-made surfaces including the large building's roof, roads, and parking areas. The spectral signature shows very low reflectance across most bands, characteristic of asphalt and dark roofing materials. \\
\hline
\textbf{E18} & 0.1526 & Soil/Dirt & Represents bare soil and dirt areas throughout the scene, particularly in open fields and around buildings. The spectral signature shows low visible reflectance with moderate NIR/SWIR reflectance, consistent with dry soil. \\
\hline
\textbf{E35} & 0.2349 & Vegetation & Represents vegetation including trees, grass, and other green areas throughout the residential zones and open spaces. The spectral signature shows characteristic vegetation features with low visible reflectance and high NIR/SWIR reflectance. \\
\hline
\textbf{E36} & 0.1113 & Roads/Asphalt & Represents paved surfaces including roads, parking lots, and driveways. The spectral signature is consistent with asphalt materials. \\
\hline
\textbf{E38} & 0.1264 & Roofing Materials & Represents various roofing materials on buildings throughout the scene. The spectral signature shows moderate reflectance across most bands, consistent with certain types of roofing materials. \\
\hline
\end{tabular}
\end{center}

\vspace{1.5ex}
\hrule
\vspace{1.5ex}

\begin{center}
    \includegraphics[width=0.18\linewidth]{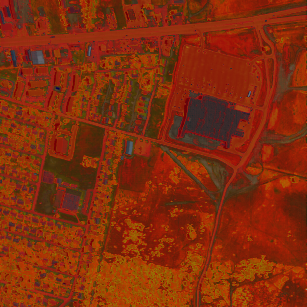} \hfill
    \includegraphics[width=0.18\linewidth]{figures/abundances/overlay_E18.png} \hfill
    \includegraphics[width=0.18\linewidth]{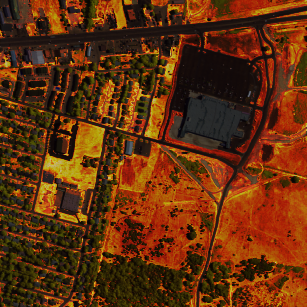} \hfill
    \includegraphics[width=0.18\linewidth]{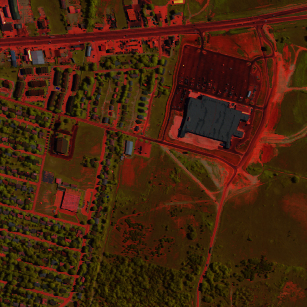} \hfill
    \includegraphics[width=0.18\linewidth]{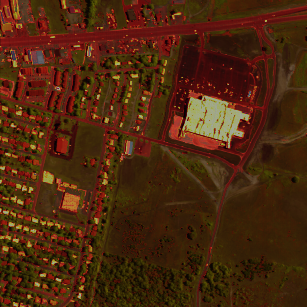}
\end{center}
\vspace{0.5ex}

\end{tcolorbox}
\caption{}
\label{fig:final_report}
\end{figure*}

Following the spatial validation, the agent queries the spectral library to form material hypotheses. Figures~\ref{fig:lib_1} and~\ref{fig:lib_2} show the library search process, demonstrating how the agent successfully grounds ambiguous spectral matches (e.g., mineral candidates and actual artificial surfaces) by cross-referencing them with the previously gathered spatial evidence. Finally, Figure~\ref{fig:final_report} presents the agent's concluding report, which summarizes the final set of active endmembers and their confidently identified physical materials.

\begin{figure*}[t]
\centering
\begin{tcolorbox}[colback=gray!5!white, colframe=black!75, title={Library Search Interaction: E11 (Dark Surfaces/Roofs)}, fonttitle=\bfseries, boxsep=0pt, left=4pt, right=4pt, top=4pt, bottom=4pt]
\scriptsize

\textbf{AI:} The abundance map for \textbf{E2} shows a strong signal on the vegetation areas, particularly in the residential zones and open fields. This appears to be similar to E23 and E6, representing vegetation. There might be redundancy between these endmembers.

Let me now perform library searches on the key endmembers to identify their materials. I'll start with the most abundant ones:
1. \textbf{E11} (dark surfaces/roofs)
2. \textbf{E23} (vegetation)
3. \textbf{E18} (soil/dirt)
4. \textbf{E10} (roads/asphalt)
5. \textbf{E3} (roofing materials)

\vspace{0.5ex}
\hrule
\vspace{0.5ex}

\textbf{[ToolMessage]} \\
\texttt{\{} \\
\texttt{~~"endmember\_id": "E11",} \\
\texttt{~~"candidates": [} \\
\texttt{~~~~\{"material": "Kieserite\_KIEDE1.a\_crse\_gr\_NIC4cc\_AREF", "category": "Minerals", ...\},} \\
\texttt{~~~~\{"material": "Lichen\_Licedea-2\_ASDFRb\_AREF", "category": "Vegetation", ...\},} \\
\texttt{~~~~\{"material": "Verdigris\_GDS789\_ASDFRa\_AREF", "category": "ArtificialMaterials", ...\},} \\
\texttt{~~~~\{"material": "Sanidine+albite\_NMNH103200\_NIC4bb\_RREF", "category": "SoilsAndMixtures", ...\}} \\
\texttt{~~~~// ... remaining candidates omitted for brevity} \\
\texttt{~~]} \\
\texttt{\}}

\vspace{0.5ex}
\hrule
\vspace{0.5ex}

\textbf{[AIMessage]} \\
The library search for \textbf{E11} shows poor matches (high spectral angles), with the closest being minerals like Kieserite and Sodium Bicarbonate. However, based on the spatial distribution (dark roofs and roads), this is likely representing dark asphalt or roofing materials. The library doesn't have good matches for these artificial materials.

\end{tcolorbox}
\caption{}
\label{fig:lib_1}
\end{figure*}

\begin{figure*}[t]
\centering
\begin{tcolorbox}[colback=gray!5!white, colframe=black!75, title={Library Search Interaction: E23 (Vegetation)}, fonttitle=\bfseries, boxsep=0pt, left=4pt, right=4pt, top=4pt, bottom=4pt]
\scriptsize

\textbf{AI:} Let me search for \textbf{E23} (vegetation):

\vspace{0.5ex}
\hrule
\vspace{0.5ex}

\textbf{[ToolMessage]} \\
\texttt{\{} \\
\texttt{~~"endmember\_id": "E23",} \\
\texttt{~~"candidates": [} \\
\texttt{~~~~\{"material": "Wollastonite\_HS348.3B\_NIC4ccc\_RREF", "category": "Minerals", ...\},} \\
\texttt{~~~~\{"material": "Lichen\_Licedea-2\_ASDFRb\_AREF", "category": "Vegetation", ...\},} \\
\texttt{~~~~\{"material": "Italian\_nat\_sienna\_GDS805\_ASDFRa\_AREF", "category": "ArtificialMaterials", ...\},} \\
\texttt{~~~~\{"material": "Sanidine+albite\_NMNH103200\_NIC4bb\_RREF", "category": "SoilsAndMixtures", ...\}} \\
\texttt{~~~~// ... remaining candidates omitted for brevity} \\
\texttt{~~]} \\
\texttt{\}}

\vspace{0.5ex}
\hrule
\vspace{0.5ex}

\textbf{[AIMessage]} \\
The library search for \textbf{E23} shows some vegetation matches (Lichen, S.altern., P.australis), but the closest matches are minerals. However, based on the spatial distribution (vegetation areas), this is clearly representing vegetation. The spectral vector shows low reflectance in visible and higher reflectance in NIR/SWIR, which is characteristic of healthy vegetation.

\end{tcolorbox}
\caption{}
\label{fig:lib_2}
\end{figure*}

\end{document}